\documentclass[review, 3p, times]{elsarticle}

\usepackage[justification=centering]{caption}

\usepackage{graphics}
\usepackage{latexsym}
\usepackage{amsfonts}
\usepackage{amssymb}
\usepackage{amsthm}
\usepackage{amsmath}
\usepackage{hyperref}
\usepackage{pdfpages}
\usepackage{tikz-qtree}
\usepackage[ruled,vlined]{algorithm2e}
\usepackage{multirow}
\usepackage{color} 
\usepackage{url}

\usepackage{amsthm}
\usepackage{IEEEtrantools}
\usepackage{mathenv}

\usepackage{listings}
\usepackage{subfigure}
\usepackage{graphics}
\usepackage{soul}
\usepackage{algpseudocode}
\usepackage{marginnote}
\usepackage{tikz-cd}
\usetikzlibrary{matrix,arrows,backgrounds}

\usepackage{url}

\usepackage{microtype}
\usepackage{graphicx}
\usepackage{float}
\usepackage{booktabs} %
\usepackage{hyperref}

\usepackage{caption}
\usepackage{natbib}

\usepackage{amsmath,amssymb,amsfonts,amsthm}

\usepackage{mathrsfs}
\usepackage{xcolor}
\usepackage{mathtools}
\usepackage{dsfont}
\usepackage{hyperref}
\usepackage{bm}
\usepackage{nicefrac}
\usepackage{wrapfig}
\usepackage{lipsum}
\usepackage{enumerate}

\let\oldReturn\Return
\renewcommand{\Return}{\State\oldReturn}

\newcounter{assumption}%
\renewcommand{\theassumption}{\arabic{assumption}}

\newcommand*\lrp[1]{\left(#1\right)}
\newcommand*\lrn[1]{\left\|#1\right\|}

\newcommand{\real}{\ensuremath{\mathbb{R}}}

\newcommand{\ve}[1]{\mathbf{#1}}

\newcommand{\x}{x}

\numberwithin{lemma}{section} 
\numberwithin{theorem}{section} 
\numberwithin{corollary}{section} 
\numberwithin{proposition}{section} 
\numberwithin{definition}{section} 
\numberwithin{example}{section} 
\numberwithin{question}{section} 

\usepackage{multirow}

\newtheoremstyle{bfnoteonly}%
{}{}%
{\itshape}{}%
{\bfseries}{.}%
{ }%
{\thmnote{#3}}

\theoremstyle{bfnoteonly}

\definecolor{brightcerulean}{rgb}{0.0, 0.53, 0.74}
\DeclareRobustCommand\full  {\tikz[baseline=-0.6ex]\draw[brightcerulean,thick] (0,0)--(0.5,0);}
\DeclareRobustCommand\fullorange{\tikz[baseline=-0.6ex]\draw[orange,thick] (0,0)--(0.5,0);}

\DeclareRobustCommand\fullblack  {\tikz[baseline=-0.6ex]\draw[black,thick] (0,0)--(0.5,0);}

\begin{document}

\title{Bayesian data-driven discovery of partial differential equations with variable coefficients}

\author{
Aoxue Chen$^{1*}$, Yifan Du$^{2*}$, Liyao Mars Gao$^{3*}$ and Guang Lin$^{4**}$}

\address{$^{1}$Department of Statistics, The University of Chicago, Chicago, IL 60637, USA\\
$^{2}$Department of Mechanical Engineering, John Hopkins University, Baltimore, MD 21218, USA\\
$^{3}$Paul G. Allen School of Computer Science \& Engineering, University of Washington, Seattle, WA 98195, USA\\
$^{4}$Department of Mathematics, Purdue University, West Lafayette, IN 47907, USA;
School of Mechanical Engineering, Purdue University, West Lafayette, IN 47907, USA}



\cortext[gl]{These authors contributed equally.}

\cortext[gl]{Corresponding author. Guang Lin, guanglin@purdue.edu}


\begin{abstract}
The discovery of Partial Differential Equations (PDEs) is an essential task for applied science and engineering. However, data-driven discovery of PDEs is generally challenging, primarily stemming from the sensitivity of the discovered equation to noise and the complexities of model selection.
In this work, we propose an advanced Bayesian sparse learning algorithm for PDE discovery with variable coefficients, predominantly when the coefficients are spatially or temporally dependent. 
Specifically, we apply threshold Bayesian group Lasso regression with a spike-and-slab prior (tBGL-SS) and leverage a Gibbs sampler for Bayesian posterior estimation of PDE coefficients.
This approach not only enhances the robustness of point estimation with valid uncertainty quantification but also relaxes the computational burden from Bayesian inference through the integration of coefficient thresholds as an approximate MCMC method. Moreover, from the quantified uncertainties, we propose a Bayesian total error bar criteria for model selection, which outperforms classic metrics including the root mean square and the Akaike information criterion. 
The capability of this method is illustrated by the discovery of several classical benchmark PDEs with spatially or temporally varying coefficients from solution data obtained from the reference simulations. In the experiments, we show that the tBGL-SS method is more robust than the baseline methods under noisy environments and provides better model selection criteria along the regularization path.
\end{abstract}


\maketitle

\section*{keywords}

Machine learning; Bayesian inference; Data-driven discovery; Bayesian group Lasso regression; Bayesian model selection; Bayesian sparse learning

\section{Introduction}
Identifying the governing laws in physical systems is of high interest in many scientific fields, including fluid dynamics, plasma dynamics, nonlinear optics, mesoscale ocean closures, computational chemistry, and so on. Many complex physical laws could be described by partial differential equations (PDEs). 
For each specific scientific task, the detailed form of the PDE often remains to be found. 
In the era of big data, when rich experimental data is available, it gives rise to an opportunity to automatically transform these data into physical laws. Recently, the fashion of data-driven discovery of hidden equations has been enabled by the rapid progress in statistics and machine learning. 

In a dynamical system taking the form of
\begin{equation}\label{intro1}
    \mathbf{u}_t = \mathbf{N} \lrp{\mathbf{u}, \mathbf{u}_x, \mathbf{u}_{xx}, \dots, x, \mu(t)},
\end{equation}
where $\mathbf{N}\lrp{\cdot}$  denotes a system of nonlinear functions of the state $\mathbf{u}(x, t)$, its spatial derivatives, and parametric dependencies through $\mu(t)$. 
We are interested in finding the expression of $\mathbf{N}\lrp{\cdot}$, which determines the general form of the equation. Many computational methods have already been put forward for this purpose. In the past few decades, symbolic regression \cite{koza1992genetic} in earlier works could uncover the structure of ordinary differential equations \cite{bongard2007automated, schmidt2009distilling}. More recently, the advancement in sparse regression methods has paved the way for a new category of approaches that view system identification as a sparse regression problem, which is applicable to both ODEs \cite{brunton2016discovering, mangan2016inferring,boninsegna2017sparse, tran2017exact, schaeffer2017bifurcation,schaeffer2017learning,schaeffer2017sparse,schaeffer2017extracting, mangan2017model,kaiser2017sparse, loiseau2018constrained} and PDEs \cite{rudy2017data, schaeffer2017learning, zhang2018robust, zhang2019robust}. Brunton et al. \cite{brunton2016discovering} propose sparse identification of dynamical systems (SINDy), which constructs a unified framework for governing equation identification. 
SINDy applies a sequential threshold least-squares algorithm, which iteratively screens out relatively small weighted coefficients. 
Similarly, PDE-find~\cite{rudy2017data} applies a sequential thresholding ridge regression algorithm to determine active terms in the PDE. 
Through enforced parsimony in SINDy (and PDE-find), the algorithm selects prominent functions in the library $\ve{\Theta}$ that rule the physical phenomenon, thereby identifying the governing equation. 
Both SINDy and PDE-find achieve huge empirical success in a wide range of applications, enabling interpretable and generalizable physics discovery. 

Even though SINDy achieves huge empirical success for many scientific applications, the assumed physical equation modeling is unable to capture PDEs with variable coefficients. In real life, the coefficients of the governing PDE typically have a strong spatial or temporal dependency. 
For example, the temperature, density, and salinity of water are likely to vary at different points spatiotemporally. This will cause a shifting effect on the fluid mechanics. 
However, the model in Eqn.~\eqref{intro1} restricts all coefficients to be constants. 
Under the spatiotemporally dependent setting, compared to the vanilla model, it is more challenging to identify the PDE with variable coefficients, as it is nontrivial to distinguish the effect of changing parameters from the dynamics. Rudy et. al. \cite{rudy2019data} employ the idea of group sparsity to enhance the PDE modeling and introduce sequentially grouped threshold ridge regression (SGTR) to capture spatial- or temporal-dependent coefficients. 
There have been many classical methods to deal with group sparsity, such as the group Lasso \cite{yuan2006model, friedman2010note}, and the Bayesian group LASSO~\cite{kyung2010penalized}. 

However, the existence of unavoidable noise brings another layer of difficulty for PDE identification with group sparsity. To be specific, there are two common sources of noise: (i) one source of noise would come from the data collection, which is constrained by the sensor resolution, data quality, and other practical difficulties; (ii) another source of noise would come from temporal differentiation for $\ve{u}_t$. The finite difference method will bring numerical instability from collected data noise with temporal discretization. 
Therefore, it is practically important to design a robust sparse regression algorithm for PDE identification. 
On top of the existence of measurement noise, it is also a challenging problem to select the correct model from the regularization path. 
As a sparse regression problem, one could always get a null model (where no variables are active) by setting a sufficiently strong sparsification effect, and always get a full model by having a sufficiently small sparsification effect. 
Introducing uncertainty into the metric of model selection could mitigate the risk of discovering an incorrect model~\citep{zhang2018robust,fung2023bayesian}. 

In this paper, we propose a robust Bayesian sparse
learning algorithm based on the Bayesian group Lasso
with spike and slab priors (BGL-SS) for PDE discovery. Using the samples drawn from the posterior distribution with a Gibbs sampler, we perform Bayesian sparse regression with uncertainty quantification, which is robust to various noise settings. 
Additionally, using the Bayesian posterior samples, we employ the Bayesian total error bar as an uncertainty-aware criterion for model selection and threshold setting. 
This Bayesian total error bar criteria avoids the collapse problem from classic first-moment-based selection criteria and improves the model selection performance in the PDE learning with variable coefficients. 
Additionally, the Bayesian total error bar criteria are also efficient to compute from the posterior MCMC samples in the previous step. 
In the experiments, we demonstrate that tBGL-SS method outperforms
 SGTR and group Lasso in noisy situations through four case studies, including linear advection equation, viscous Burger's equation, advection-diffusion Equation, and Kuramoto-Sivashinsky Equation. 
 We examine and visualize the uncertainty estimation from tBGL-SS in the linear advection equation in Sec.~\ref{sec:Uncertainty quantification}. 
 For model selection criteria, we compare the Bayesian total error bar criteria to the mean-squared error and an Akaike information criterion inspired loss (AIC-loss). 
 In Sec.\ref{sec:Comparison of Model Selection Criteria}, the Bayesian total error bar clearly shows a better regularization path for model selection.  
Finally, we discuss three noise filters with the robust Bayesian sparse learning algorithm, aiming to reach better results for practitioners when dealing with very large noise. 

The main contribution of this work is three-fold:
\begin{itemize}
    \item We propose a Bayesian sparse learning algorithm based on threshold Bayesian group Lasso with spike and slab prior (tBGL-SS) to identify the partial differential equations with temporal or spatial varying coefficients. Our algorithm can robustly identify governing equations under noisy environments and is computationally efficient for PDE discovery. 
    \item We enable valid uncertainty quantification from the Bayesian modeling, and we utilize the uncertainty information to compute the Bayesian total error bar criteria for model selection, which performs model selection with minimal uncertainty under noisy environments and outperforms other model selection methods. 
    \item We implement the algorithm and conduct numerical experiments to study the effect of tBGL-SS. Experimental results show strong improvements of tBGL-SS in PDE discovery under various cases. We further test different preprocessing methods of noise filters to deal with the PDE discovery under large observation noise, which could robustify the learning outcomes. 
\end{itemize}

The paper is organized as follows. In Section 2, the general ideas of the proposed data-driven learning problem is described. The main methodology is discussed in Section 3. The case studies are demonstrated in Section 4 to illustrate the capability of the proposed methods. We present the conclusion in Section 5.

\section{Background}
\subsection{Data-driven discovery of time-dependent PDE}
\label{gi}
We first review the case where the PDE has constant coefficients. SINDy algorithm proposed a general framework that utilizes sparse regression for model discovery~\citep{brunton2016discovering} and PDE-FIND enables PDE discovery via Group LASSO~\citep{rudy2019data}. Take (\ref{intro1}) as an example. We first select a set of candidate common functions containing derivatives with different orders. For example, we could set up the library as follows: 
\begin{equation}\label{gi0.1}
  \ve{\Theta}(\ve{u})=\left\{1, \ve{u}, \ve{u}_x, \ve{u}^{2}, \ve{u} \ve{u}_x, \ve{u}_x^2, \dots, \ve{u}_{xx}^2\right\}. 
\end{equation}

We adopt the framework of equation discovery in \cite{brunton2016discovering,rudy2017data,rudy2019data} as follows. We have the evolution of PDE data $\ve{U}=\left\{\ve{u}(x_i,t_j)\right\}_{i,j}^{n,m}$ where $x_{1:n}$ represents spatial points and $t_{1:m}$ represents time points. We define $\boldsymbol{\xi}=[\xi_1, \xi_2, \dots, \xi_p]^T\in\real^p$ as a vector of all coefficients. 
it turns into a linear regression problem:
\begin{equation}\label{gi1}
\left(\begin{array}{c}{\ve{u}_{t}\left(x_{1}, t_{1}\right)} \\ {\ve{u}_{t}\left(x_{2}, t_{1}\right)} \\ {\vdots} \\ {\ve{u}_{t}\left(x_{n}, t_{m}\right)}\end{array}\right)=\left(\begin{array}{cccccccc}{1} & {\ve{u}\left(x_{1}, t_{1}\right)} & {\ve{u}_{x}\left(x_{1}, t_{1}\right)} & {\ve{u}^2\left(x_{1}, t_{1}\right)} & {\ve{u}\ve{u}_x\left(x_{1}, t_{1}\right)} & {\ve{u}_{x}^2\left(x_{1}, t_{1}\right)} & \dots
              & {\ve{u}_{xx}^2\left(x_{1}, t_{1}\right)} \\
{1} & {\ve{u}\left(x_{2}, t_{1}\right)} & {\ve{u}_{x}\left(x_{2}, t_{1}\right)} & {\ve{u}^2\left(x_{2}, t_{1}\right)} & {\ve{u}\ve{u}_x\left(x_{2}, t_{1}\right)} & {\ve{u}_{x}^2\left(x_{2}, t_{1}\right)}
              & \dots
              & {\ve{u}_{xx}^2\left(x_{2}, t_{1}\right)} \\ {\vdots} & {\vdots} & {\vdots} & {\vdots} & {\vdots} & {\vdots} & & {\vdots} \\
{1} & {\ve{u}\left(x_{n}, t_{m}\right)} & {\ve{u}_{x}\left(x_{n}, t_{m}\right)} & {\ve{u}^2\left(x_{n}, t_{m}\right)} & {\ve{u}\ve{u}_x\left(x_{n}, t_{m}\right)} & {\ve{u}_{x}^2\left(x_{n}, t_{m}\right)} 
& \dots
& {\ve{u}_{xx}^2\left(x_{n}, t_{m}\right)}\end{array}\right) \boldsymbol{\xi}
\end{equation}

The problem above can be cast into linear equation solving with $\ve{A}\ve{x}=\ve{b}$. In many cases, due to the existence of measurement noise and missing data, we alternatively solve an optimization problem that minimizes the loss function

\begin{equation}
    \boldsymbol{\xi} = \arg\min_{\boldsymbol{\xi}\in\real^p} \lrn{\ve{U}-\Theta\lrp{\ve{U}}\boldsymbol{\xi}}_2^2. 
\end{equation} 

In Statistics, the estimation of $\boldsymbol{\xi}$ is called regression. Parsimonious modeling (which enforces sparsity in $\boldsymbol{\xi}$) enables governing equation identification from the library terms given $\ve{U}$.

When the parameters of PDE are time-dependent, these parameters should be alternatively considered as separate time series. Therefore, the formulation above should be rewritten as
\begin{equation}\label{gi2.1}
  \ve{u}_{t}=\left(1, \ve{u}, \ve{u}_x, \ve{u}^2, \ve{u}\ve{u}_x,\ve{u}_x^2\right)\boldsymbol{\xi}(t), 
\end{equation}
where $\boldsymbol{\xi}(t)=[\xi_{const}(t), \xi_{\ve{u}}(t), \xi_{\ve{u}_x}(t), \xi_{\ve{u}^2}(t), \xi_{\ve{u}\ve{u}_x}(t), \xi_{\ve{u}_x^2}(t)]^T$. Every $\xi_i(t)$ denotes a time series that represents the time evolution of $\xi_i$. 

To handle this, it is proper to begin with solving the equation at each time step. The regression at each time step looks like
\begin{equation}\label{gi2.2}
  \left(\begin{array}{c}{\ve{u}_{t}\left(x_{1}, t_{i}\right)} \\ {\ve{u}_{t}\left(x_{2}, t_{i}\right)} \\ {\vdots} \\ {\ve{u}_{t}\left(x_{n}, t_{i}\right)}\end{array}\right)
  =\left(\begin{array}{ccccccc}{1} & {\ve{u}\left(x_{1}, t_{i}\right)} & {\ve{u}_{x}\left(x_{1}, t_{i}\right)} & \dots & {\ve{u}_{x}^2\left(x_{1}, t_{i}\right)} & \dots
              & {\ve{u}_{xx}^2\left(x_{1}, t_{i}\right)} \\
{1} & {\ve{u}\left(x_{2}, t_{i}\right)} & {\ve{u}_{x}\left(x_{2}, t_{i}\right)} & \dots & {\ve{u}_{x}^2\left(x_{2}, t_{i}\right)} & \dots
              & {\ve{u}_{xx}^2\left(x_{2}, t_{i}\right)} \\
{\vdots} & {\vdots} & {\vdots} & & \vdots & & \vdots \\
{1} & {\ve{u}\left(x_{1}, t_{i}\right)} & {\ve{u}_{x}\left(x_{n}, t_{i}\right)} & \dots & {\ve{u}_{x}^2\left(x_{n}, t_{i}\right)} & \dots
              & {\ve{u}_{xx}^2\left(x_{n}, t_{i}\right)} \\\end{array}\right) \boldsymbol{\xi}(t_{i}), \quad i=1,2,\dots ,m
\end{equation}
which every $t_i$ can be denoted as
\begin{equation}\label{gi2.3}
  \ve{u}_t^{(i)}=\boldsymbol \Theta(u^{(i)})\boldsymbol{\xi}^{(i)}, \quad i=1,2,\dots ,m
\end{equation}

By incorporating sparsity constraints, since every $\xi^{(i)}$ should share the same sparsity pattern, the regression problem above  can be integrated into a single system as described in~\cite{rudy2019data}: 
\begin{equation}\label{gi3.1}
  \left(\begin{array}{c}{\ve{u}_{t}^{(1)}} \\ {\ve{u}_{t}^{(2)}} \\ {\vdots} \\ {\ve{u}_{t}^{(m)}}\end{array}\right)=\left(\begin{array}{cccc}{\boldsymbol \Theta\left(\ve{u}^{(1)}\right)} & {} & {} & {} \\ {} & {\boldsymbol \Theta\left(\ve{u}^{(2)}\right)} & {} & {} \\ {} & {} & {\ddots} & {} \\ {} & {} & {} & {\boldsymbol \Theta\left(\ve{u}^{(m)}\right)}\end{array}\right)\left(\begin{array}{c}{\boldsymbol{\xi}^{(1)}} \\ {\boldsymbol{\xi}^{(2)}} \\ {\vdots} \\ {\boldsymbol{\xi}^{(m)}}\end{array}\right)
\end{equation}

Equivalently, we have the matrix form
\begin{equation}
    \ve{U}=\boldsymbol \Theta \boldsymbol{\xi}.
\end{equation}

\paragraph{Bayesian sparse regression}

In this system, we put these coefficients into different groups. For example, the first term in every $\boldsymbol{\xi^{(i)}}$ will form a group of variables, corresponding to the time series from the PDE parameter. 
If a candidate term does not present in this PDE, the coefficients in this group should be zero uniformly. 
This falls into the setup of regression under group sparsity, which could be addressed by methods like the group Lasso \cite{yuan2006model} and SGTR \cite{rudy2019data}. 
Among the extant methods, Bayesian group Lasso can provide a distributional approximation, which is hard for non-Bayesian methods. Exact zero estimates can be obtained using the posterior median in BGL-SS to facilitate variable selection. The details for BGL-SS will be discussed in the following section.
\section{Methods}
\label{methods}
\subsection{Bayesian formulation in PDE Discovery with spatiotemporal-dependent coefficients}
We consider the following grouped sparse linear regression model for spatiotemporal-dependent PDE discovery: 
\begin{equation}
    \mathbf{u}_t = \sum_{g=1}^G \ve{\Theta}_g(\ve{u})\boldsymbol\xi_g+\mathbf{\epsilon},
\end{equation}
where $\ve{\Theta}_g(\ve{u})$ stand for the $g$-th term in the library, $\ve{\xi}_g = [\ve{\xi}_g^{(1)}, \ve{\xi}_g^{(2)}, ..., \ve{\xi}_g^{(m)}]^T$, $\boldsymbol\epsilon\sim\mathcal{N}(0,\sigma^2 \boldsymbol{I}_{n\times n})$, and $p$ denotes the total number of predictors where $p=mG$. Notice that in our framework of PDE discovery, the grouped variables must have the same cardinality. 

In the Bayesian setting, the model parameter $\boldsymbol{\xi}$ is considered as a random variable. Then, we utilize Bayesian inference algorithms to obtain the distributional estimate of $\boldsymbol{\xi}$ from the observation data $(\boldsymbol{\Theta}, \boldsymbol{u})$ and a given prior $\pi(\boldsymbol{\xi})$. To achieve group sparsity, it is common to employ Bayesian group Lasso algorithm. In specific, we set the prior of $\boldsymbol{\xi}_g$ that $\pi\left(\boldsymbol{\xi}_{g}\right) \propto \exp \left(-\frac{\lambda}{\sigma}\left\|\boldsymbol{\xi}_{g}\right\|\right)$,
where $\sigma,\lambda\in\mathbb{R}$. This prior could be interpreted as a Gamma mixture of normals \cite{kyung2010penalized}. In specific, the hierarchical model of the Bayesian group Lasso is shown as follows:
\begin{equation}
\label{eqn:naive_group_lasso}
\begin{array}{c}
\boldsymbol{U} \mid \boldsymbol{\Theta}, \boldsymbol{\xi}, \sigma^{2} \sim \boldsymbol{N}_{n}\left(\boldsymbol{\Theta} \boldsymbol{\xi}, \sigma^{2} \boldsymbol{I}_{n}\right) \\  \boldsymbol{\xi}_{g} \mid \sigma^{2}, \tau_{g}^{2} \stackrel{i n d}{\sim} N_{m}\left(\mathbf{0}, \sigma^{2} \tau_{g}^{2} \boldsymbol{I}_{m}\right) \\ \tau_{g}^{2} \stackrel{i n d}{\sim} \operatorname{Gamma}\left(\frac{m+1}{2}, \frac{\lambda^{2}}{2}\right), \quad g\in \mathcal{G} 
\end{array}
\end{equation}
where $\mathcal{G}$ denotes the collection of groups. 

However, the above Bayesian group Lasso setting cannot produce exact zero results. 
In other words, all posterior samples are not sparse. 
In the PDE discovery setting, it is required to have enforced sparse components since the sparsity pattern represents the corresponding governing equation. 
To resolve this problem, we incorporate the spike-and-slab prior~\cite{xu2015bayesian} using an independent zero-inflated mixture prior to each regression parameter to ensure sparsity. The improved hierarchical model is expressed as
\begin{equation}
\begin{array}{c}
\boldsymbol{U} \mid \boldsymbol{\Theta}, \boldsymbol{\xi}, \sigma^{2} \sim \boldsymbol{N}_{n}\left(\boldsymbol{\Theta} \boldsymbol{\xi}, \sigma^{2} \boldsymbol{I}_{n}\right) \\
\boldsymbol{\xi}_{g} | \sigma^{2}, \tau_{g}^{2} \stackrel{i n d}{\sim}\left(1-\pi_{0}\right) \boldsymbol{N}_{m}\left(\mathbf{0}, \sigma^{2} \tau_{g}^{2} \boldsymbol{I}_{m}\right)+\pi_{0} \delta_{0}\left(\boldsymbol{\beta}_{g}\right) \\
{\tau_{g}^{2} \stackrel{i n d}{\sim} \text { Gamma }\left(\frac{m+1}{2}, \frac{\lambda^{2}}{2}\right), \quad g \in \mathcal{G}} \\
{\sigma^{2} \sim \text { Inverse Gamma }(\alpha, \gamma), \quad \sigma^{2}>0}\end{array}
\end{equation}
where $\delta_{0}\left(\boldsymbol{\xi}_{g}\right)$ is a point mass at zero. The value of $\pi_0$ and $\lambda$ can be regarded as tunable hyperparameters. Or alternatively, it is also possible to follow the empirical Bayes framework and estimate $\pi_0$ and $\lambda$ directly from data. 

\paragraph{Posterior sampling via block Gibbs sampler.} To sample this hierarchical Bayesian model, we utilize a block Gibbs sampler to the posterior distribution $p\left(\boldsymbol{\xi}, \boldsymbol{\tau}^{2}, \sigma^{2}, \pi_{0} \mid \boldsymbol{U}, \boldsymbol{\Theta}\right)$. We write the full posterior distribution in the following
\begin{align}
    p\lrp{\boldsymbol{\xi}, \tau^2, \sigma^2, \pi_0\mid\boldsymbol{u},\boldsymbol{\Theta}} &\propto (\sigma^2)^{-\frac{n}{2}} \exp\left\{-\frac{1}{2\sigma^2}\lrp{\boldsymbol{u}-\boldsymbol{\Theta}\boldsymbol{\xi}}^T\lrp{\boldsymbol{u}-\boldsymbol{\Theta}\boldsymbol{\xi}}\right\}\nonumber\\
    &\times \prod_{g=1}^G \left[(1-\pi_0)(2\pi\sigma^2\tau_g^2)^{-\frac{m}{2}}\exp\left\{-\frac{\boldsymbol{\xi}_g^T\boldsymbol{\xi}_g}{2\sigma^2\tau_g}\right\}\mathbf{1}(\boldsymbol{\xi}_g\neq 0) + \pi_0\delta_0(\boldsymbol{\xi}_g)\right] \\
    &\times \prod_{g=1}^G \lambda^{m+1}\tau_g^{m-1}\exp\lrp{-\frac{\lambda^2}{2}\tau_g^2} \times \pi_0^{a-1}(1-\pi_0)^{b-1}\times (\sigma^2)^{-\alpha-1}\exp\lrp{-\frac{\gamma}{\sigma^2}}\nonumber,
\end{align}
where $\mathbf{1}\lrp{\cdot}$ denotes the indicator function. 

The block Gibbs sampler~\cite{hobert1998geometric} efficiently simulates the posterior distribution $p\left(\boldsymbol{\xi}, \boldsymbol{\tau}^{2}, \sigma^{2}, \pi_{0} \mid \boldsymbol{U}, \boldsymbol{\Theta}\right)$. Let $\boldsymbol{\xi}_{(g)}=\lrp{\xi_1^T, \xi_2^T, ..., \xi_{g-1}^T, \xi_{g}^T, ..., \xi_{G}^T}^T$ and $\boldsymbol{\Theta}_{(g)} = \lrp{\boldsymbol{\Theta}_1^T, \boldsymbol{\Theta}_2^T, ..., \boldsymbol{\Theta}_{g-1}^T, \boldsymbol{\Theta}_{g}^T, ..., \boldsymbol{\Theta}_{G}^T}^T$. Since we assume a Gaussian linear model for $\boldsymbol{U}\mid \boldsymbol{\Theta}, \boldsymbol{\xi}, \sigma^2$, we generate $\boldsymbol{\xi}_g\mid\text{rest}$ via
\begin{align}
    \boldsymbol{\xi}_g\mid\text{rest} \sim(1-l_g) \mathcal{N}\lrp{\boldsymbol{\mu}_g, \sigma^2\boldsymbol{\Sigma}_g} + l_g\delta_0\lrp{\boldsymbol{\xi}_g}, \;\;\;\;g=1,...,G,
\end{align}
where
\begin{align}
    \boldsymbol{\mu}_g=\boldsymbol{\Sigma}_g\boldsymbol{\Theta}_g^T\lrp{\boldsymbol{U}-\boldsymbol{\Theta}_{(g)}\boldsymbol{\xi}_{(g)}},\boldsymbol{\Sigma}_g=\lrp{\boldsymbol{\Theta}_g^T\boldsymbol{\Theta}_g+\frac{1}{\tau_g^2}\boldsymbol{I}_m}^{-1}
\end{align}
and, from the Bayes formula, $l_g=p(\boldsymbol{\xi}_g=0\mid\text{rest})$ is 
\begin{align}
    l_g=\frac{\pi_0}{\pi_0+(1-\pi_0)\tau_g^{-m}|\boldsymbol{\Sigma}_g|^{\frac{1}{2}}\exp\left\{\frac{1}{2\sigma^2}\lrn{\boldsymbol{\Sigma}_g^{\frac{1}{2}} \boldsymbol{\Theta}_g^T \lrp{\boldsymbol{U}-\boldsymbol{\Theta}_{(g)}\boldsymbol{\xi}_{(g)}}}_2^2\right\}}.
\end{align}
For other variables in the posterior distribution, $\sigma^2$ is generated from a inverse Gamma distribution, $\pi_0$ is generated from a Beta distribution, and the inverse of $\tau_g^2$ is generated from a mixture of inverse Gamma and inverse Gaussian distribution~\cite{xu2015bayesian}. By generating posterior samples of $\boldsymbol{\xi}, \tau^2, \sigma^2, \pi_0$, we are able to perform Bayesian inference on $p\left(\boldsymbol{\xi}, \boldsymbol{\tau}^{2}, \sigma^{2}, \pi_{0} \mid \boldsymbol{U}, \boldsymbol{\Theta}\right)$. 

\subsection{Thresholding Bayesian group Lasso with the spike-and-slab prior (tBGL-SS)}

However, Bayesian MCMC is computationally demanding. If we naively apply BGL-SS algorithm to the PDE finding task with varying coefficients, 
the temporal and spatial dependencies will tremendously enlarge the number of variables, which makes the algorithm practically infeasible.
For example, dealing with a system with 256 time steps and 256 spatial sample points, the baseline BGL-SS algorithm takes over an hour to find the varying coefficients. 
Additionally, without longer MCMC runs and hyperparameter specification, the BGL-SS algorithm cannot
reach a better performance, and fails to exclude variables with coefficients close to
zero.

To resolve the above-mentioned challenges, we use the idea of approximate MCMC and incorporate thresholding procedures in the following. 
In the context of sparse model discovery, thresholding-based procedures have been successful in many scientific tasks~\citep{rudy2017data, rudy2019data, zhang2018robust}. Incorporating thresholds is also a viable solution from the Bayesian perspective. From the idea of approximate MCMC~\cite{johndrow2020scalable}, it includes a hard-thresholding procedure to improve per-iteration computational complexity without sacrificing accuracy. When the target solution is sparse, we expect a large subset of the variables to have a small posteriori, and hence
thresholding the entries smaller than a small threshold should not affect the accuracy of the
algorithm. Removing these redundant variables from the library has significant computational advantages for MCMC algorithms.
Jointly inspired by the idea of thresholding-based physics discovery and approximate MCMC, we propose our algorithm of threshold Bayesian group Lasso with spike-and-slab prior in the following. 

\begin{algorithm}[H]
\SetAlgoLined
\caption{Threshold Bayesian Group Lasso with Spike and Slab Prior: $\boldsymbol{u}=\boldsymbol{\Theta}\boldsymbol{\ve{\xi}}$}
\label{algorithm_tbglss}
\KwIn{$\boldsymbol{u}$, $\boldsymbol{\Theta}$, $\mathcal{G}$, threshold $c_\textit{threshold}$, thresholding criterion $f(\boldsymbol{\xi}_g)$}
\KwOut{$\tilde{\boldsymbol{\xi}}$, $s_i^2$, $i=1,2,\dots \sum_{g=1}^G m_g$}
\While{True}{

Simulate the posterior distribution $p(\boldsymbol{\xi}|\boldsymbol{u},\boldsymbol{\Theta})$ in $\boldsymbol{u}=\boldsymbol{\Theta}\boldsymbol{\xi}$ using the \textit{block Gibbs sampler} in BGL-SS

Use the sample median $\tilde{\boldsymbol{\xi}}$ as the estimated value of coefficients, and use the sample variance $s_i^2$ as the estimated variance of each coefficient

For all $\tilde{\boldsymbol{\xi}_g}$ inferior to the threshold ($f(\boldsymbol{\xi}_g)<c_\textit{threshold}$), remove them from the system

\If{No group is removed}{
break\;
}

\If{No group is left}{
break\;
}

Delete the columns of $\boldsymbol{X}$ corresponding to the deleted groups\;
}
\label{alg:tbgl-ss}
\end{algorithm}

From the posterior samples, it is possible to remove inappropriate groups of terms from the regression system. 
In Alg.~\ref{alg:tbgl-ss}, we consider a statistics $f(\Tilde{\boldsymbol{\xi}})$ and compare it with a threshold $t$. Considering sparsity at the group level, for the choice of $f(\cdot)$, the posterior mean of the coefficients is not a good choice, as some significant coefficients may perturbate around zero within the group. Alternatively, we consider the following criteria. 

\paragraph{Confine the scale of coefficient}
The first one confines the scale of each coefficient. If the sparsity is at the individual level, a threshold of absolute value can be used~\cite{zhang2018robust}. The norm of coefficients in each group can correctly indicate the scale of each varying coefficient, thus is applicable in the PDE finding task. The maximum of the absolute values in each group may also be applicable in some specific cases. In this paper, the root mean square within each group is used as the example of scale thresholds:
\begin{equation}\label{rms}
    f(\boldsymbol{\xi}_g)=\frac{\left\|\boldsymbol{\xi}_{g}\right\|_2}{\sqrt{m}}
\end{equation}

It must be noted that the value of thresholds must be carefully tuned to get an accurate result. If the threshold is too low, the algorithm is not capable of deleting insignificant terms and causes overfitting. If the threshold is too high, some real terms may also be screened out. In some specific cases, the correct coefficients may have either smaller scales or greater standard deviations, and thus cannot be detected when the threshold is set high if you only use one criterion of the threshold.


\paragraph{Consideration in computation} Alg.~\ref{alg:tbgl-ss} iteratively removes groups of variables from the posterior samples. The $\text { Total Iterations }=\text { Iterations per Update } \times \text { Number of Updates }$. To minimize the computational cost, the total iterations should be as few as possible. 
In practice, we notice that the block Gibbs sampler does not need as many iterations in each update as in the original algorithm. As an approximate MCMC method, we only require the first several updates for candidate reduction for a rough estimate of the sparsity pattern. 
The computational complexity of the block Gibbs sampler reduces rapidly with decreasing size of the library. 
Additionally, in the experiments, the sparsity pattern estimate converges quickly with a few repetitions. Therefore, we only have to repeat the Gibbs sampler for BGL-SS with a few rounds. In short, unlike many other Bayesian inference methods, Alg.~\ref{alg:tbgl-ss} is tractable and computationally efficient.

\subsection{Model selection and Bayesian evidence}
Some parameters such as the value of thresholds must be set before running the algorithm. To find the best setting, the algorithm shall be tested using many different parameter values and different criteria for model evaluation.

When it comes to the PDE identification task, the mean square error of coefficients is a proper measurement of how close is the fitted model to the true partial differential equation. However, it cannot be used for model selection as we do not know anything about the true equation in practice.

An important criterion for model evaluation is an Akaike information criterion (AIC)-inspired loss function \cite{mangan2017model, rudy2019data}:
\begin{equation}\label{aic}
    \mathcal{L}(\boldsymbol{\xi})=N \ln \left(\frac{\left\|\tilde{\boldsymbol{\Theta}} \boldsymbol{\xi}-\tilde{\boldsymbol{u}}\right\|_{2}^{2}}{N}+\epsilon\right)+2 k.
\end{equation}
Here $k$ equals the number of coefficients that are not zeros in the fitted equation, $\tilde{\boldsymbol{\Theta}}$ and $\tilde{\boldsymbol{u}}$ are the normalized form of the matrix $\boldsymbol{\Theta}$ and $\boldsymbol{u}$ in $\boldsymbol{u}=\boldsymbol{\Theta}\boldsymbol{\xi}$, and $N$ is the size of $\boldsymbol \Theta$. The term $\epsilon$ is a small value used to avoid overfitting. A lower loss hints a model has a better fit to data.

A huge drawback of the AIC-inspired loss is that AIC ignores the uncertainty as pushing the sample to the asymptotic. In the Bayesian framework, the Bayesian evidence is a golden rule for model selection which is formally defined as
\begin{align}
    \mathcal{Z} = p\lrp{\boldsymbol{u} \mid \mathcal{M}} &= \int p\lrp{\boldsymbol{u}\mid\boldsymbol{\xi}} p\lrp{\boldsymbol{\xi}\mid\mathcal{M}}d\boldsymbol{\xi}
\end{align},
where $\mathcal{M}$ denotes model $\mathcal{M}$ including the data generating model, sequentially thresholded library, and hyperparameter setting. Via posterior samples, one could approximate the Bayesian evidence by~\cite{newton1994approximate}
\begin{align}
    \mathcal{Z} = \frac{1}{\int \frac{\pi(\ve{\xi}\mid \ve{u}, \mathcal{M})}{p(\ve{u}\mid\ve{\xi}, \mathcal{M})}d\ve{\xi}}.
\end{align}

We alternatively consider $\frac{1}{\mathcal{Z}}$ to let a smaller value denotes a better model. Using posterior samples, this quantity can be empirically estimated via
\begin{align}
    \frac{1}{\hat{\mathcal{Z}}} = \sum_{i=1} \frac{1}{p(\ve{u}\mid\ve{\xi}_i, \mathcal{M})},\;\;\;\;\xi_i\sim\pi(\xi\mid\ve{u}, \mathcal{M}).
\end{align}

The spirit of the above metric is to jointly consider the posterior likelihood and uncertainty. On the one hand, for equation discovery, it is easy to falsely identify terms with similar local behavior. 
For instance, since the Taylor expansion of $\sin(z)$ is $\sin(z)\approx z-\frac{1}{6}z^3$, all physical models with $\sin(z)$ are likely to be identified as a linear combination of  $\sin(z, )z, z^3$. 
Take a moving pendulum as an example. The physics of a moving pendulum could be expressed by model $\mathcal{M}_A$ with only $\sin(z)$, but the local physics could also be perfectly described by another model $\mathcal{M}_B$ with $\sin(z), z, z^3$. 
One cannot distinguish these two models purely via the likelihood, but we can see the difference when we include the uncertainty. Every posterior sample tends to have smaller $p(\boldsymbol{u}\mid\boldsymbol{\xi}_i, \mathcal{M}_B)$ which will cause $\frac{1}{\hat{\mathcal{Z}_B}}$ to be large. 

On the other hand, when a model $\mathcal{M}'$ falsely excludes a correct library term, the posterior samples tend to be very uncertain, which aims to capture the unexplained effect. Additionally, all $p(\boldsymbol{u}\mid\boldsymbol{\xi}_i, \mathcal{M}')$ is expected to be small with the missing covariate, leading large $\frac{1}{\hat{\mathcal{Z}_B}}$. Evaluating the above formulation may lead to challenges in computation with strong assumptions on the data-generating model. Therefore, in the context of physics discovery, we construct a total "error bar" metric as follows inspired by \cite{zhang2018robust}:
\begin{equation}\label{te}
    \mathcal{E}(\boldsymbol \xi)=\sum_{g \in \mathcal{G}'\atop \left\|\boldsymbol{\xi}_{g}\right\|_2\neq0}\sum_{i=1}^{m}\frac{s_{g,i}^2}{\left\|\boldsymbol{\xi}_{g}\right\|_2^2}
\end{equation}
The set $\mathcal{G}'\subseteq \mathcal{G}$ is the collection of remaining groups. $\mathcal{E}(\boldsymbol{\xi})$ only requires statistics that have been previously computed in Algorithm 1. The total error bar metric similarly considers the uncertainty of a model and promotes large $\lrn{\boldsymbol{\xi}_g}_2^2$ to avoid Taylor expanded terms. $\mathcal{E}(\boldsymbol{\xi})$ is also expected to be large when the library term is missing from the current model, as it will significantly increase the uncertainty $s_{g,i}$. 
These criteria could help to evaluate candidate models via different perspectives. We compare these selection metrics in Section \ref{nr}.

\section{Numerical Results} \label{nr}

\textcolor{black}{In this section, the capabilities and strengths of the proposed tBGL-SS algorithm are demonstrated through numerical experiments. In Section \ref{sec:Comparison of Algorithm}, the tBGL-SS algorithm is applied to several PDE problems to identify the unknown spatiotemporal profile of unknown coefficients. This section serves as the presentation of the most general results and a benchmark against other methods. The following Sections \ref{sec:Uncertainty quantification}, \ref{sec:Comparison of Model Selection Criteria}, \ref{sec:largenoise} are detailed and specific analyses regarding the uncertainty quantification, model selection, and treatment of large noises for the tBGL-SS algorithm. }

\subsection{Comparison of Algorithms}
\label{sec:Comparison of Algorithm}

\textcolor{black}{We examine the accuracy of the tBGL-SS algorithm on four PDE models: linear advection equation, viscous Burgers equation with temporally dependent coefficients, advection-diffusion equation with spatial dependency, and Kuramoto-Sivashinsky equation with spatial dependency. These PDE models are selected due to their respective mathematical properties, which potentially impact the performance of the proposed algorithm. The data sets are obtained from reference numerical simulations with pre-determined initial and boundary conditions. The PDE problems are first transformed into an ODE system through appropriate spatial discretizations, then solved by calling the 'lsode' integrator through scipy interface. The tBGL-SS method is applied to these data sets to reconstruct the form of the dynamic equations and the corresponding unknown coefficients. The performance of tBGL-SS method is compared with baseline methods SGTR \cite{rudy2019data} and group Lasso \cite{yuan2006model}. To quantify the effect of data uncertainty on the reconstructed PDE, we evaluate the accuracy of the reconstructed solution from data sets injected with white noises of different levels.}

\subsubsection{Linear advection equation}

\begin{figure}[H]\centering
  \includegraphics[width=1.0\textwidth]{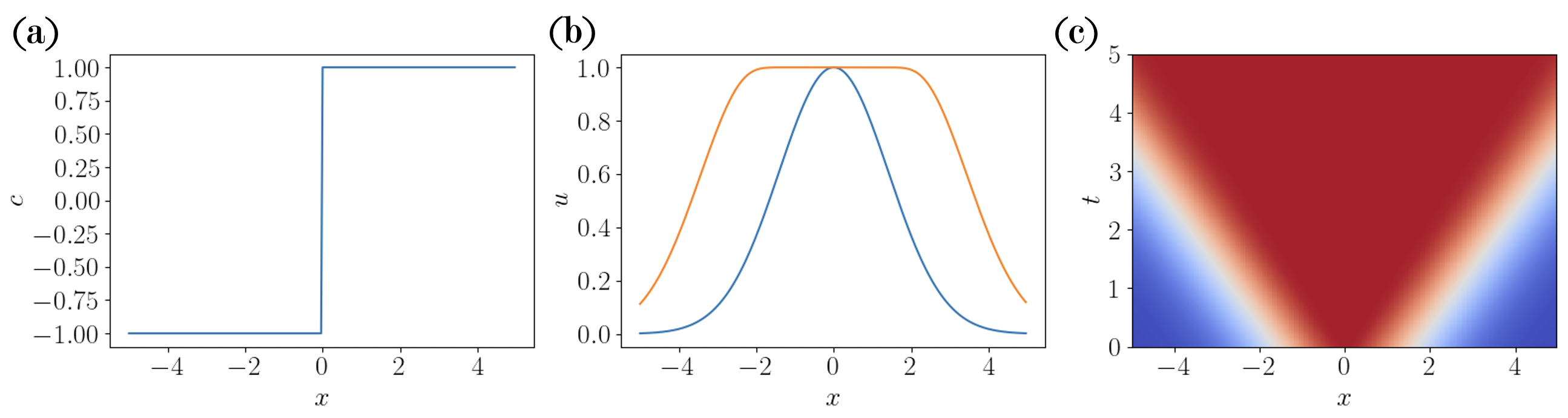}
  \caption{Coefficient and solution of in-homogeneous advection equation. (a): waves speed. (b): solution of wave equation at two different times. \full: $t=0$; \fullorange: $t=2$. (c): spatiotemporal contour of solution $u$. } \label{fig: advection solution}
\end{figure}

\begin{figure}[H]\centering
  \includegraphics[width=1.0\textwidth]{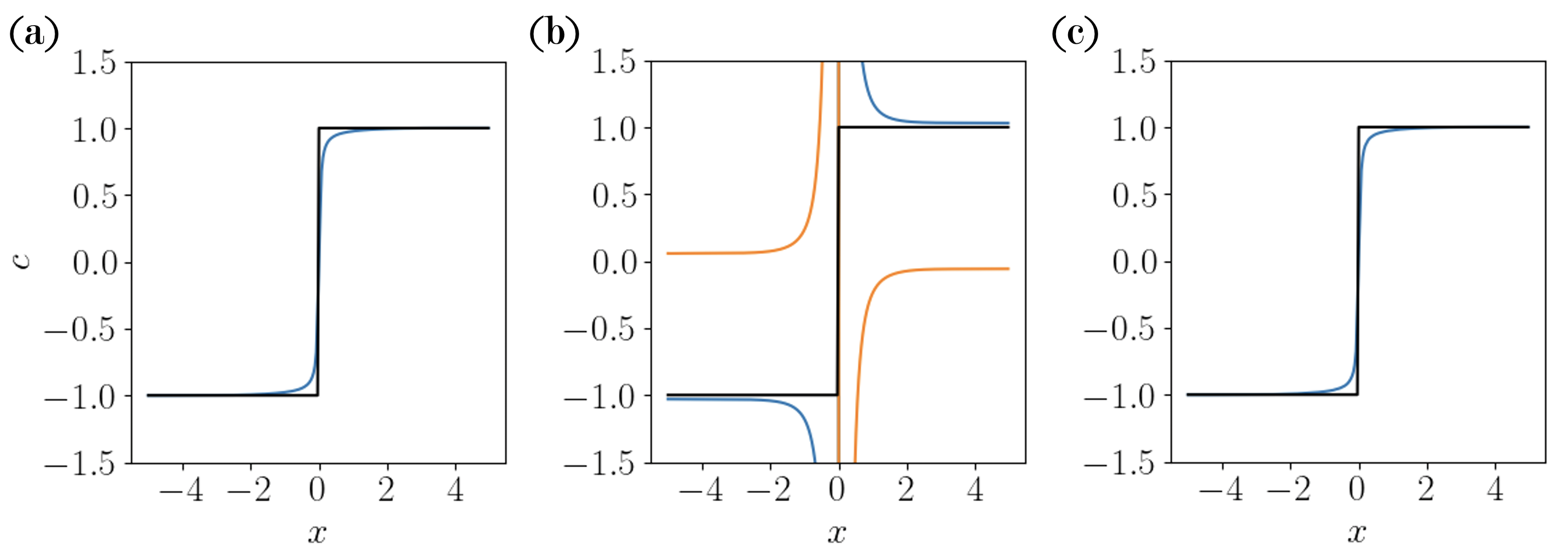}
  \caption{The equation coefficients learned from data by different algorithms. (a) SGTR, (b) group lasso, (c) tBGLSS. \fullblack: True solution of wave speed, \full: discovered wave speed from data.\fullorange: false coefficient of second-order spatial derivative discovered from the data. } \label{fig: advection coefficients}
\end{figure}

We demonstrate an elementary PDE - the linear advection equation - to exemplify the PDE system with spatio-temporal dependency.  In this case study, we aim to offer a general glimpse of the tBGL-SS and the baseline methods from the prior work~\citep{rudy2019data}.
The linear advection equation describes the propagation of a wave inside certain media. We consider the 1-D advection equation of the following form: 
\begin{equation}
    \frac{\partial u}{\partial t} + c(x)\frac{\partial u}{\partial x}=0. 
\end{equation}
The solution $u(x,t)$ models the motion of a wave with a speed $c(x)$ inside a spatially in-homogeneous media. The solution to this simple equation could exhibit complex behavior when the wave speed depends on spatial coordinates. In particular, we consider a piece-wise constant wave speed: 
\begin{equation}
   c(x) = 
\left\{
    \begin{array}{lr}
        -1, & \text{if } x<0\\
        1, & \text{if } x>0, 
    \end{array}
\right.
\end{equation}
and an initial condition of Gaussian shape: 
\begin{equation}
    u(x,0)=\exp[-x^2]. 
\end{equation}
The true wave speed and numerical solution of this equation are visualized in Figure \ref{fig: advection solution}. The true wave speed is negative and positive at $x<0$ and $x>0$, which transports the initial condition toward the left and the right direction in these two regions. This PDE problem is spatially discretized using an upwind scheme and integrated forward in time to obtain the full-time history of the solution. The resulting solution at $t>0$ expands outwards and forms a plateau of $u=1$ around $x=0$. The numerical solution of this PDE problem is utilized to discover the original PDE model. This problem serves as a demonstration of the capability of our method for identifying in-homogeneous pure advection effects. The equation terms and coefficients discovered from the data using different algorithms are visualized in Figure \ref{fig: advection coefficients}. The SGTR and tBGL-SS methods both identify the first-order derivative term $c\partial u/\partial x$ correctly and discover the spatial-dependent wave speed accurately except near $x=0$. The group lasso method not only finds the first-order derivative advection term from the data but also falsely discovers a second-order spatial derivative term, which corresponds to the diffusion effect. The coefficients determined by group lasso approach to correct value away from $x=0$, but reach very large spikes of order $O(500)$ near $x=0$. There are two possible reasons why the reconstruction accuracy deteriorates at this location. First, the discontinuity of the wave speed poses additional difficulty for the discovery of the equation coefficient. Second, the data around $x=0$ is not informative for the reconstruction of wave speed, since the first time and spatial derivative of solution $u$ are both nearly zero at this location. The data provided in this problem does not provide enough information to fully determine the wave speed at $x=0$, which affects all three algorithms. Although this problem is fundamentally difficult to address without more data, quantification of uncertainty provides credibility for the reconstructed coefficients, which benefits from the Bayesian statistical framework adopted in this work. The uncertainty quantification of this problem is discussed in detail in Section \ref{sec:Uncertainty quantification}. 

\subsubsection{Burger's Equation}
Let us consider a typical example of PDE. In Burger's equation
\begin{equation} \label{burgers-1}
    \frac{\partial u}{\partial t}+\mu u \frac{\partial u}{\partial x}=\nu \frac{\partial^{2} u}{\partial x^{2}}
\end{equation}
$\mu$ is set to be a time-dependent oscillating coefficient, and $\nu$ to be constant throughout time. Using \textit{SciPy} function \textit{odeint}, the equation can be numerically solved. We use the tool functions from \cite{rudy2019data} and follow their way of calculating derivatives and handling the equation. Twenty candidate terms in the partial differential equation, with powers of $u$ up to third order multiplying derivatives of $u$ with respect to $x$ up to fourth order, are considered.

\begin{figure}[H]\centering
  \includegraphics[width=14cm]{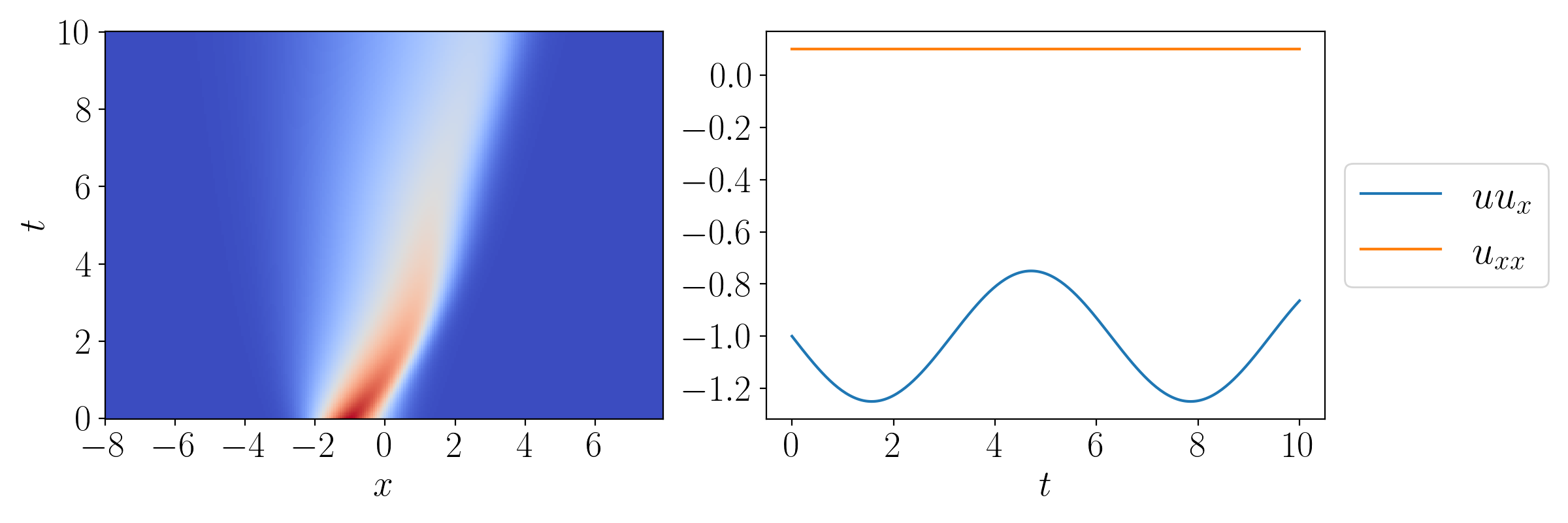}
  \caption{Left: Numeral solution of Burger's equation, with initial conditions stated in (\ref{burgers-2}). Right: Time-varying coefficients in Burger's equation.} \label{nr1.1}
\end{figure}

Set the coefficients and the initial condition:
\begin{equation}\label{burgers-2}
\begin{array}{c}
    \mu=1+\sin(t)/4,\quad\nu=0.1 \\
    u(x,0)=\exp[-(x+1)^2]
\end{array}
\end{equation}
and we get the numeral solutions as shown in Figure \ref{nr1.1}.

\begin{figure}[H]\centering
  \includegraphics[width=14cm]{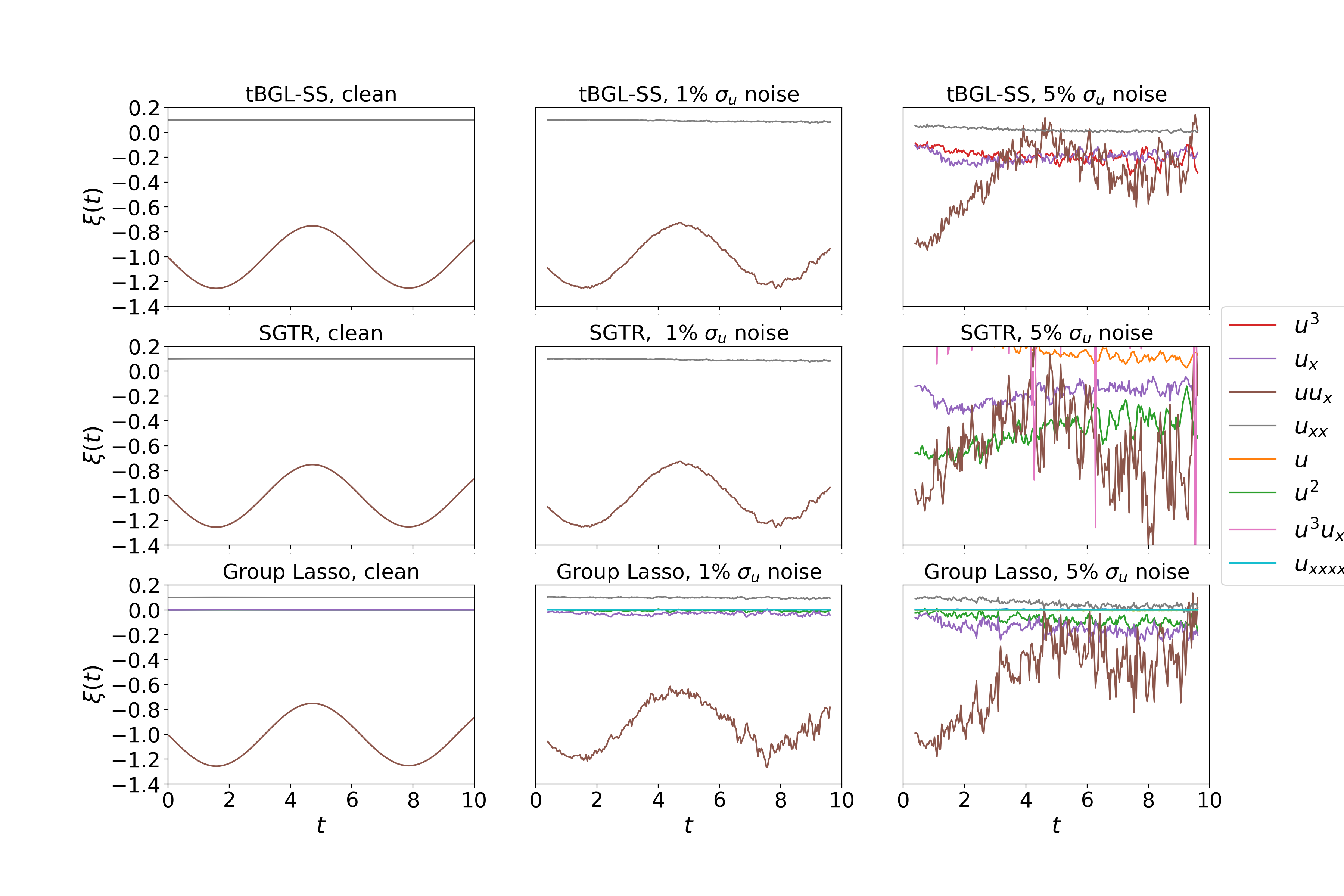}
  \caption{Nine experiments conducted on data with three different levels of noise, Left: Clean, Middle: 1\% $\sigma_u$ noise, Right: 5\% $\sigma_u$ noise using three methods, Top: tBGL-SS, Middle: SGTR, and Bottom: Group Lasso in learning Burger's equation.} \label{nr1.2}
\end{figure}

The numeral solutions are then used to generate derivatives and construct the linear system in (\ref{gi2.2}). Threshold BGL-SS is used along with SGTR and Group Lasso to solve the same regression. The results are shown in Figure \ref{nr1.2}.

Three algorithms are tested with three different levels of white noise added to the data. 
Upon the clean data and the 1\% $\sigma_u$ noise data, a $t_{RMS}=0.02$ is used. When the noise is risen to 5\%, $t_{RMS}$ changes to $0.01$. The $t_{GE}=0.1$ in all three experiments. The $\lambda$ of the Lasso is calculated using Monte Carlo EM algorithm \cite{park2008bayesian, xu2015bayesian}. For SGTR and group Lasso, several solutions with different settings of parameters are generated, and the model with the lowest AIC-like loss is chosen.

When the noise level is low or none, both tBGL-SS and SGTR yield good results, while group Lasso failed to exclude some coefficients constantly close to zero, and the output is least fit to the true model. But when the noise is at a relatively high level, all three algorithms fail to identify the terms and coefficient values of the embedded partial differential equation. Nevertheless, threshold BGL-SS and group Lasso appear to be more robust for larger noise than SGTR.

\begin{figure}[H]\centering
  \includegraphics[width=10cm]{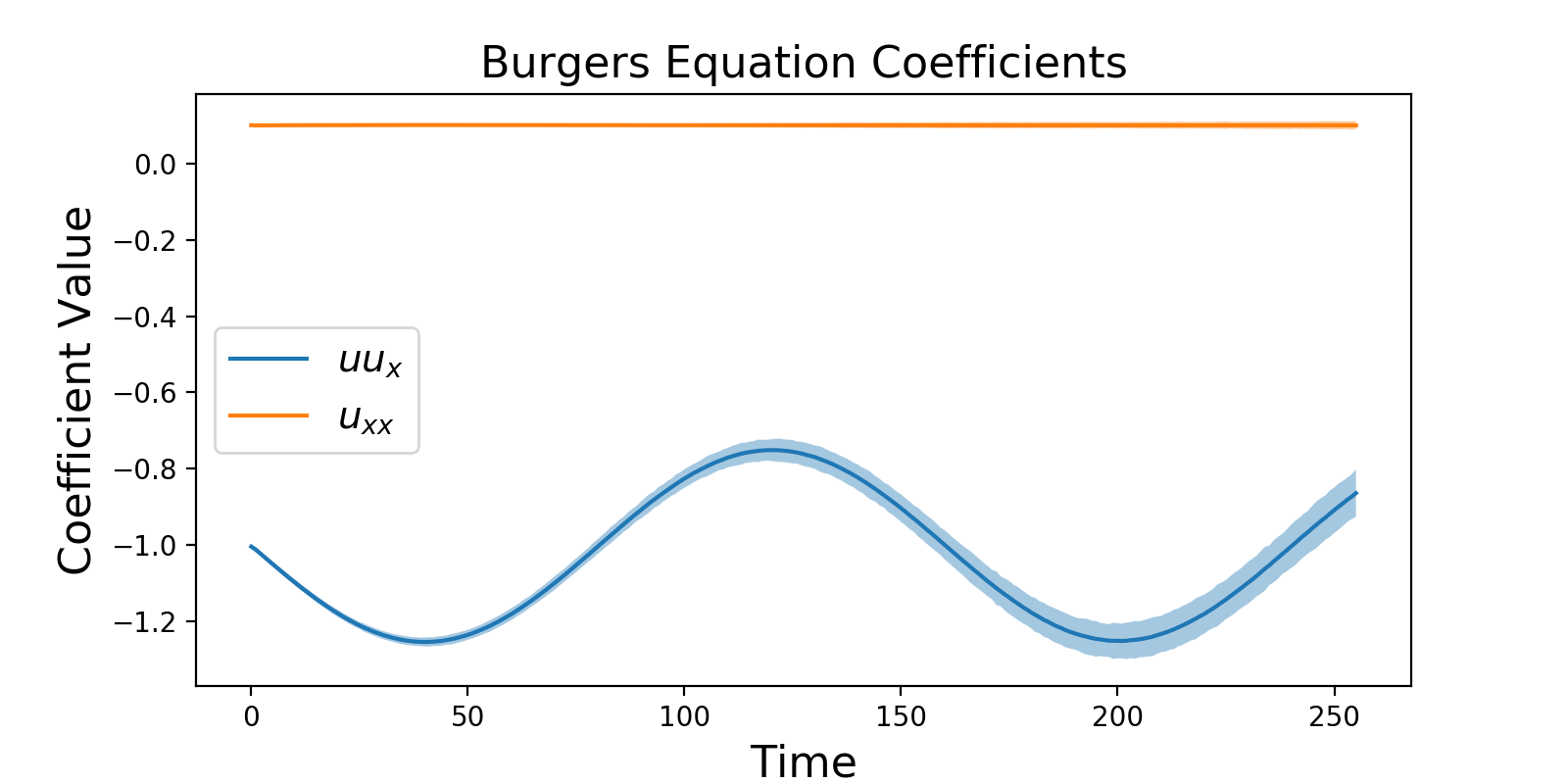}
  \caption{The shaded area represents the standard deviations of coefficients at each moment. Any points in the area is within one SD from the median. } \label{nr1.3}
\end{figure}

An advantage of tBGL-SS is that the error bars (or bands) can easily be constructed. An example of sample standard deviation error bands is illustrated in Figure \ref{nr1.3}. 

\subsubsection{Advection-Diffusion Equation}
\begin{figure}[H]\centering
  \includegraphics[width=14cm]{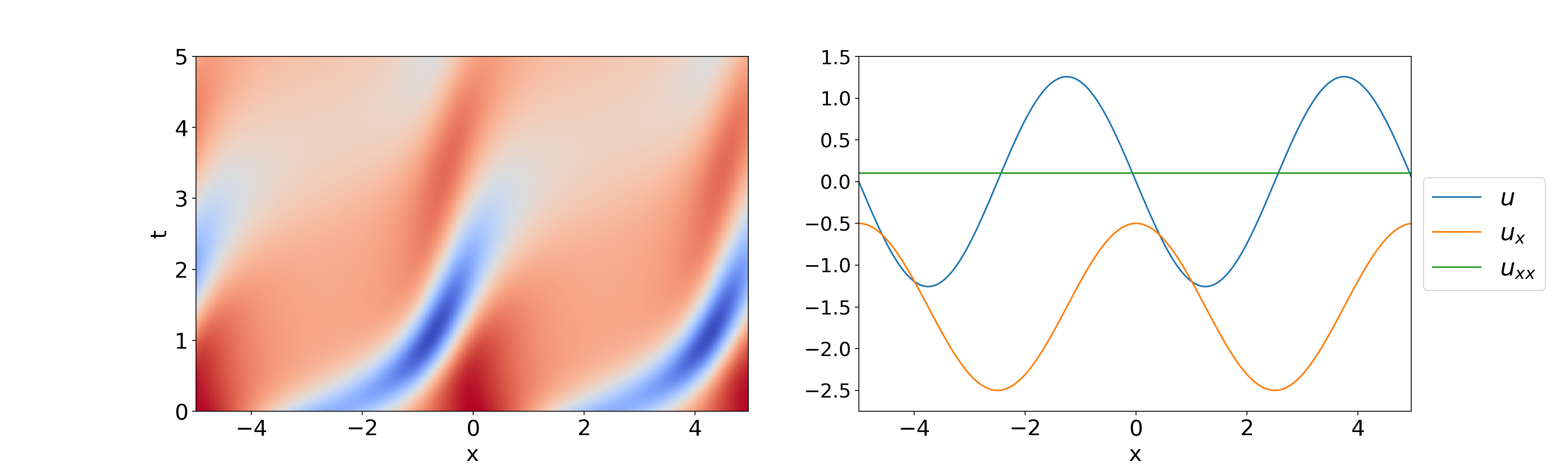}
  \caption{Left: Numeral solution of a spatially dependent Advection-Diffusion Equation, with initial conditions stated in (\ref{ad-2}). Right: Spatially dependent coefficients in the Advection-Diffusion equation.} \label{nr1.4}
\end{figure}

Our second example aims to test whether tBGL-SS is able to identify spatially dependent coefficients. The Advection-Diffusion equation is used:
\begin{equation}[H]\label{ad-1}
    \frac{\partial u}{\partial t}=\frac{\partial(\mu u)}{\partial x}+\nu \frac{\partial^2u}{\partial x^2}=\frac{\partial\mu}{\partial x}u+\mu\frac{\partial u}{\partial x}+\nu \frac{\partial^2u}{\partial x^2}
\end{equation}
Here we let $\mu$ be a spatially varying coefficient. The values of coefficients and the initial condition is set as follows:
\begin{equation}\label{ad-2}
    \begin{array}{c}
        \mu=-1.5+\cos (0.4 \pi x),\quad \nu=0.1\\
        u(0)=\cos(0.4\pi x)
    \end{array}
\end{equation}
We solve the partial differential equation within the spatial range $[-5,5]$ from $t=0$ to $5$ as shown in Figure \ref{nr1.4}.

\begin{figure}[H]\centering
  \includegraphics[width=14cm]{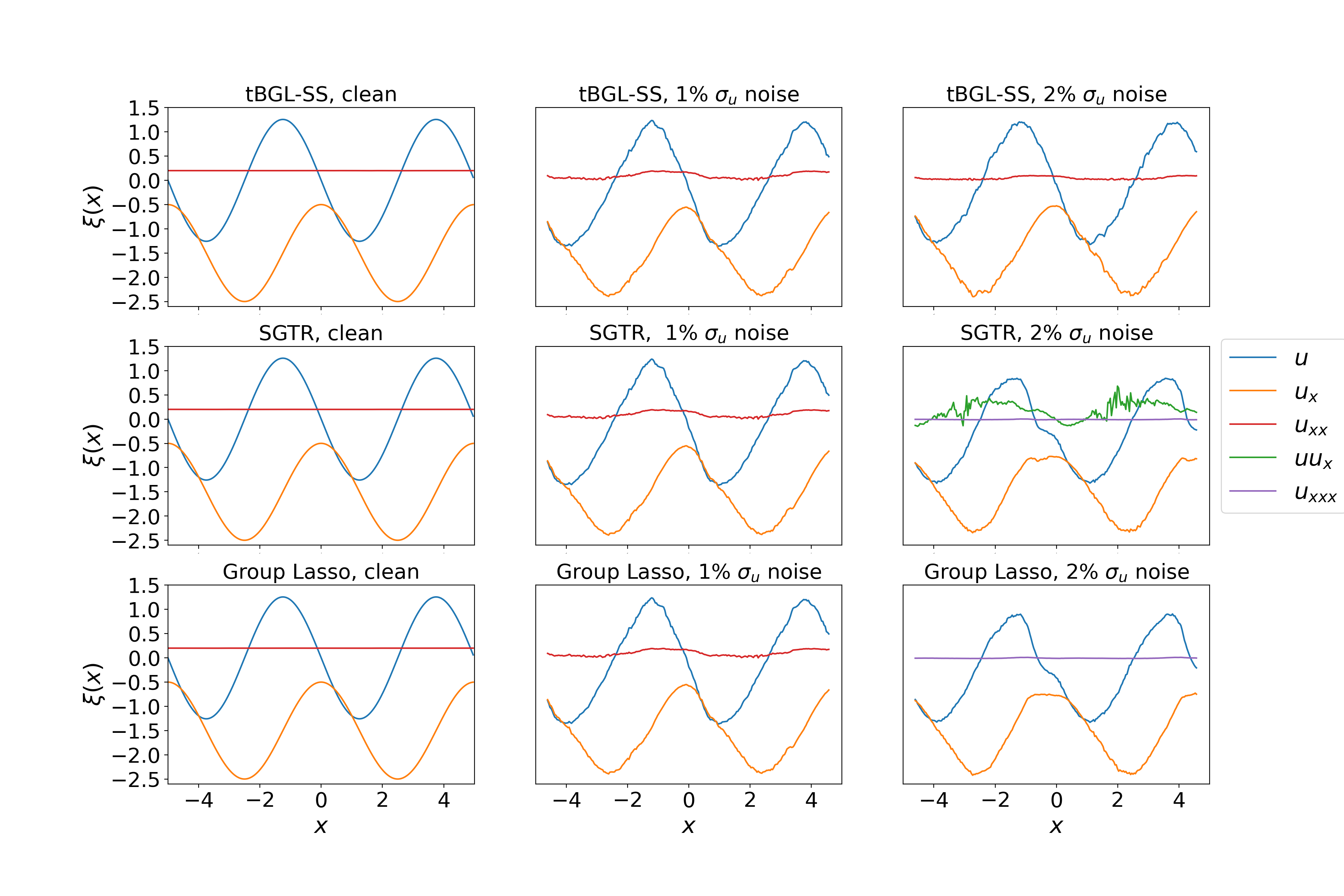}
  \caption{Nine experiments conducted on data with three different levels of noise, Left: Clean, Middle: 1\% $\sigma_u$ noise, Right: 2\% $\sigma_u$ noise using three methods, Top: tBGL-SS, Middle: SGTR, Bottom: Group Lasso in learning Advection-Diffusion equation.} \label{nr1.5}
\end{figure}

We still use threshold BGL-SS, SGTR, and Group Lasso respectively, on three noise levels. The results are presented in Figure \ref{nr1.5}. For threshold BGL-SS, a $t_{RMS}=0.02$ is applied on the clean data and the 1\% $\sigma_u$ noise data, and $t_{RMS}=0.01$ for the 2\% $ \sigma_u$ noise data. The $t_{GE}=0.08$ in all three experiments. More iterations in the Gibbs sampler are used in the 2\% noise situation. For SGTR and group Lasso, the best parameters, corresponding to the model with the lowest AIC-like loss, is chosen automatically.
The three methods work equally accurately when the data is clean and with small noise. But when larger noise is added, threshold BGL-SS seems to work better than SGTR and Group Lasso. Both Group Lasso and SGTR failed to include the term $u_{xx}$ but instead identified other false terms.

\subsubsection{Kuramoto-Sivashinsky Equation}
\begin{figure}[H]\centering
  \includegraphics[width=14cm]{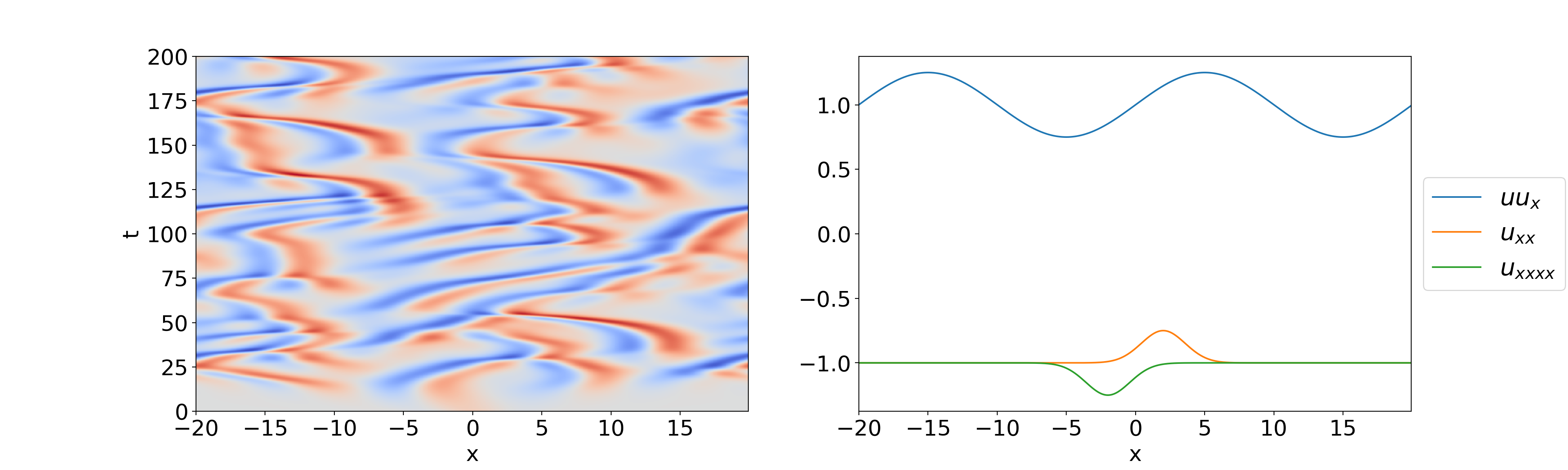}
  \caption{Left: Numeral solution of KS Equation, with initial conditions stated in (\ref{KS-2}). Right: Spatially dependent coefficients in the KS equation.} \label{nr1.6}
\end{figure}

\begin{figure}[H]\centering
  \includegraphics[width=14cm]{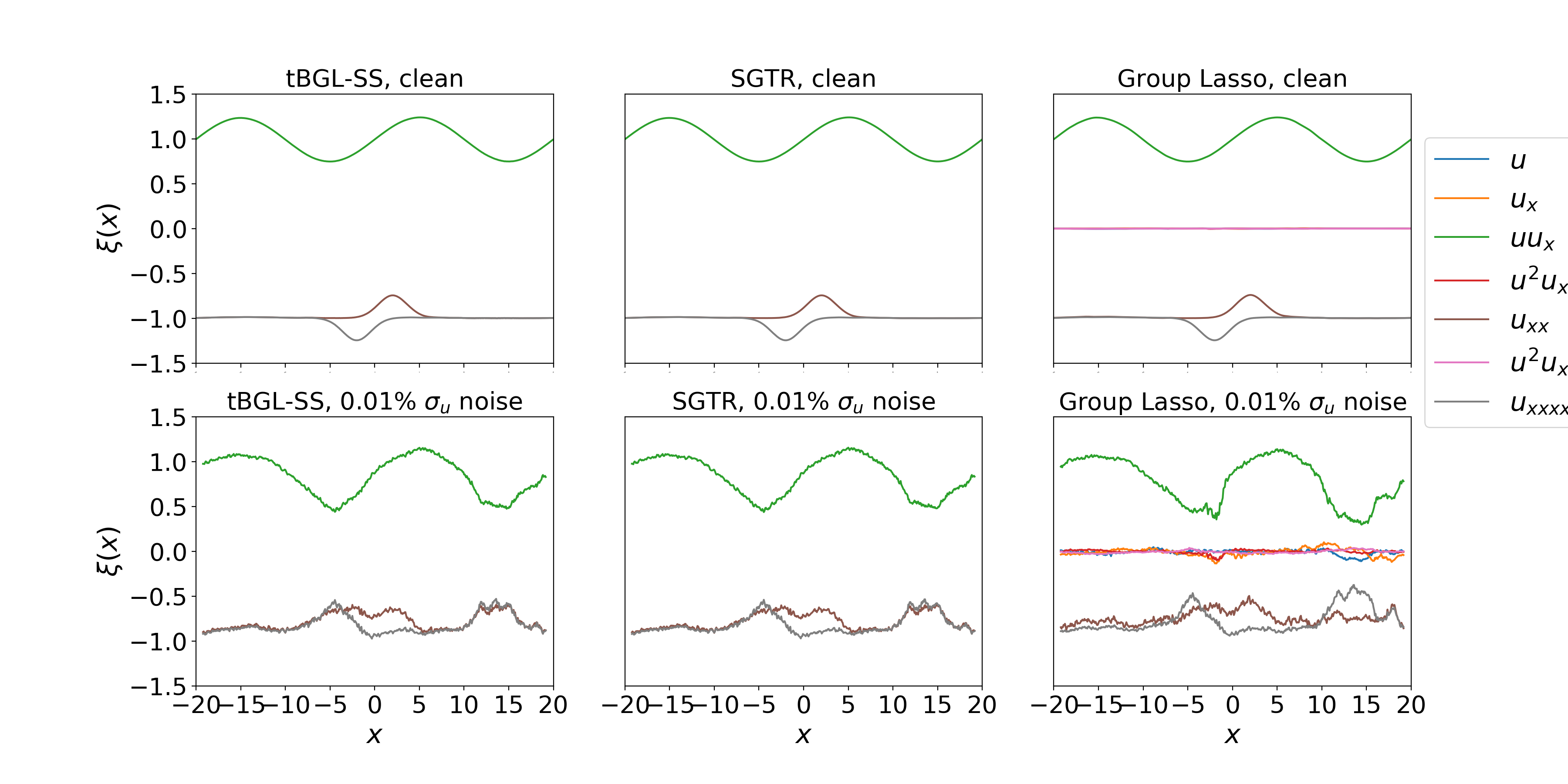}
  \caption{Six experiments conducted on data, Top: with noise, Bottom: without noise, using three methods, Left: tBGL-SS, Middle: SGTR, and Right: Group Lasso in learning KS equation.} \label{nr1.7}
\end{figure}

The third test example we use is a Kuramoto-Sivashinsky Equation:
\begin{equation}\label{KS-1}
    \frac{\partial u}{\partial t}=\alpha u\frac{\partial u}{\partial x}+\beta \frac{\partial^2 u}{\partial x^2}+\gamma\frac{\partial^4 u}{\partial x^4}
\end{equation}
Here we let
\begin{equation}\label{KS-2}
\begin{array}{c}
    \alpha=1+0.25\sin(0.1\pi x)\\ \beta=-1+0.25e^{-\frac{(x-2)^2}{5}}\\ \gamma=-1-0.25e^{-\frac{(x+2)^2}{5}}\\
    u(0)=e^{-x^2}
\end{array}
\end{equation}
and get the solution within the spatial range $[-20,20]$ from $t=0$ to $200$ as in Figure \ref{nr1.6}.
The KS equation tends to have a chaotic solution, and we use the grid points from $t=100$ to consider the more chaotic part only. When there is 1\% noise, none of the three methods can yield a correct result, due to the presence of a fourth-order derivative in the equation. Instead, the 0.01\% noise is used. For threshold BGL-SS, a $t_{RMS}=0.1$ and a $t_{GE}=0.05$ are applied. The results are shown in Figure \ref{nr1.7}.
\subsection{Uncertainty quantification}
\label{sec:Uncertainty quantification}
\begin{figure}[H]\centering
  \includegraphics[width=0.7\textwidth]{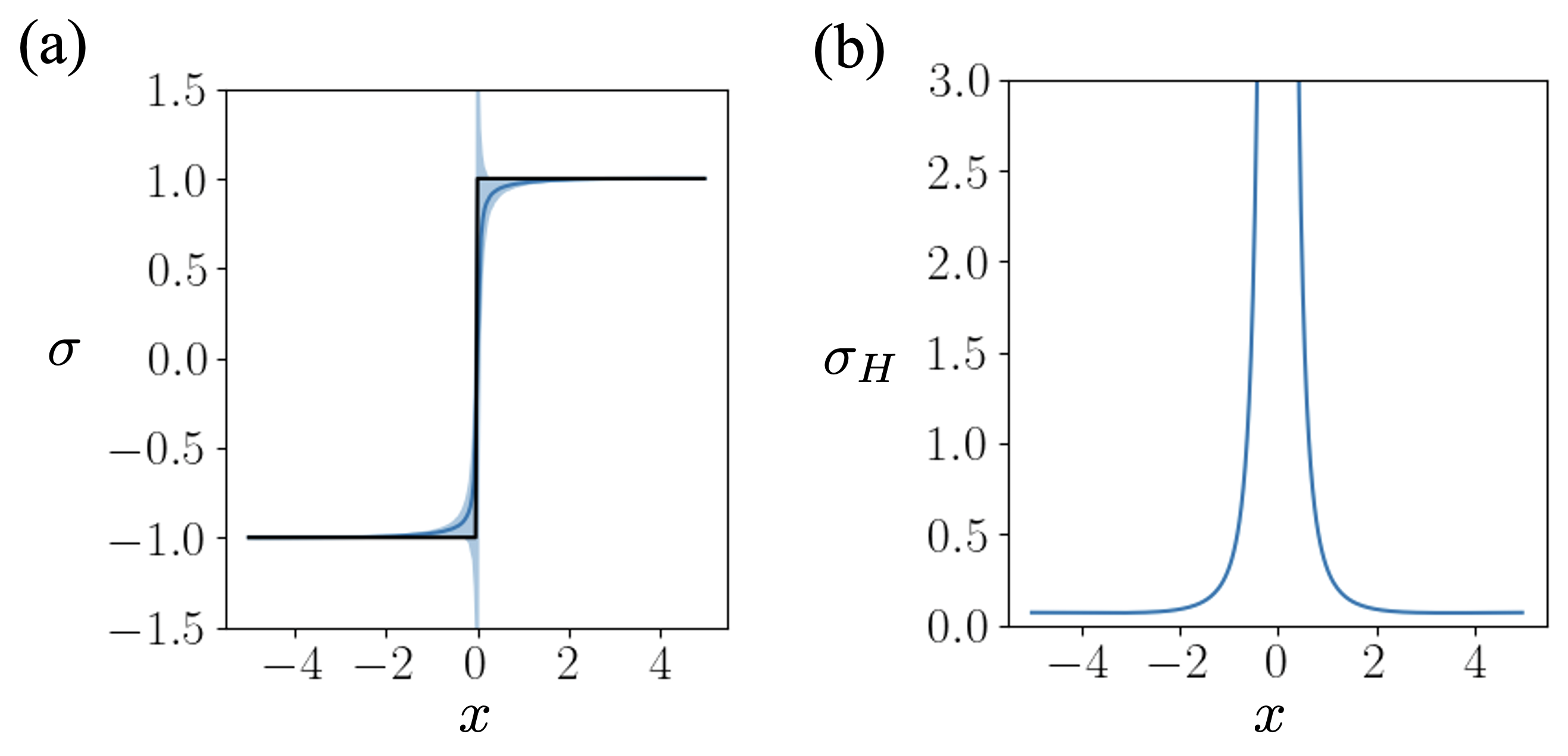}
  \caption{Uncertainty estimations. } \label{fig:UQ}
\end{figure}

The Bayesian framework adopted in this work naturally enables uncertainty quantification for the discovered coefficients. In this section, we revisit the example of the linear advection equation in Section \ref{sec:Comparison of Algorithm}, and present the uncertainty quantification of the reconstructed wave speed. As mentioned in Section \ref{sec:Comparison of Algorithm}, there is an inherent difficulty in the identification of the advection equation given the solution data in Figure \ref{fig: advection solution}(c) near $x = 0$. Near this point, the derivatives $\partial u/\partial x$ and $\partial u/\partial t$ are both zero at all times, thus the data do not provide sufficient information for the identification of the equation coefficient at this point. Using this example, we demonstrate that our method tBGLSS is capable of identifying the location where the discovery of equation coefficients is inaccurate and fundamentally difficult without the knowledge of a true solution. 
The tBGLSS solution is most inaccurate near the discontinuity at $x=0$ where the error bar is also the widest. This indicates that standard deviation from Gibbs samples can effectively find the spatial region where the solution contains large uncertainty which originates from the algorithm and ill-posedness of the problem. The uncertainty estimated from our method can be compared with other measures of uncertainty. For example, the difficulty, or the ill-posedness of the regression problem can also be measured by the Hessian matrix of the $L^2$ loss: 
\begin{equation}
    \mathcal{J}=\frac{1}{2}\lrn{\ve{U}-\Theta\lrp{\ve{U}}\boldsymbol{\xi}}_2^2, \,\,\mathbf{H}=\nabla^2\mathcal{J}=\Theta^{\top}\Theta. 
\end{equation}
The $i$-th diagonal element of $\mathbf{H}$ represents the sensitivity of loss $\mathcal{J}$ with respect to the coefficient at location $x_i$. The inverse of diagonal elements, $\sigma_{H}(x_i)=1/H_{ii}$, can also be used to evaluate the uncertainty of coefficients. For example, a nearly zero value of $H_{i,i}$ (a very large value of $\sigma_H$) indicates that the coefficient $\boldsymbol{\xi}$ at $x_i$ makes no impact on the value of loss function $\mathcal{J}$, thus a very large uncertainty for the equation coefficient. In Figure \ref{fig:UQ}(b) we visualize the profile of $\sigma_H$ depending on $x$, which gives a qualitatively similar estimation of uncertainty as the tBGLSS method. 
\subsection{Comparison of Model Selection Criteria}
\label{sec:Comparison of Model Selection Criteria}
It is critical to perform model selection from the path of regularization. 
For real-world applications, randomly deciding the threshold will lead to incorrect model inference. 
A feasible way of choosing the right threshold values is to perform a regularization path study, which tests the model's performance with increasing sparsity constraints. 
From a collection of models generated from these increasingly strong settings, one could pick the optimal model based on different evaluation metrics. 
In this subsection, we investigate the three different criteria of model selection: the mean squared error of coefficients, the AIC-like loss (\ref{aic}), and the Bayesian total error bar (\ref{te}). Note here that the mean squared error of coefficients cannot work in practice since the true coefficient is unknown.
We include this metric here as a hypothetical criterion for model selection. 

\begin{figure}[H]
  \begin{minipage}{0.33\linewidth}\centering
    \includegraphics[width=5.2cm]{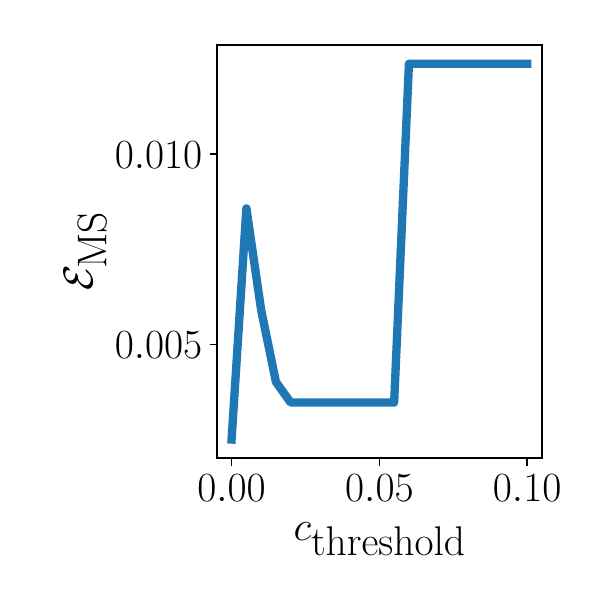}
  \end{minipage}
  \begin{minipage}{0.33\linewidth}\centering
    \includegraphics[width=5.2cm]{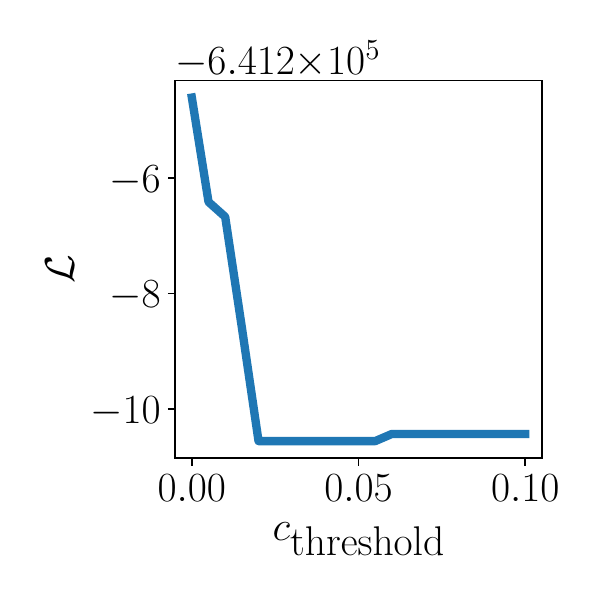}
  \end{minipage}
  \begin{minipage}{0.33\linewidth}\centering
    \includegraphics[width=5.2cm]{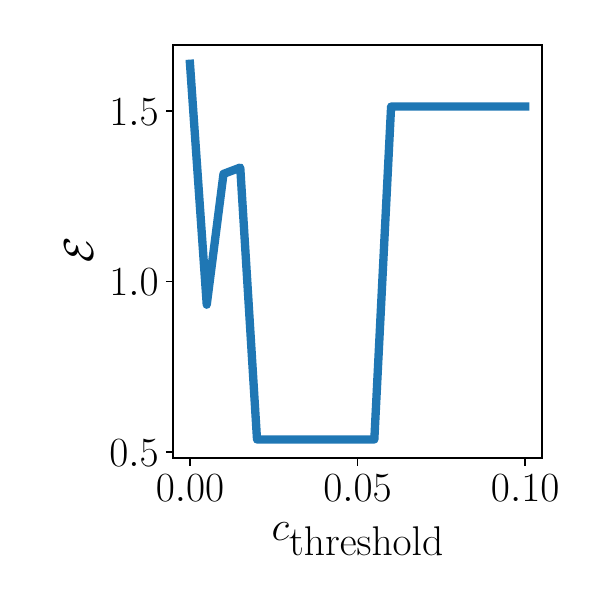}
  \end{minipage}
  \caption{Three criteria of model selection, Left: Mean squared error based criterion, Middle: AIC-like loss based criterion, Right: Bayesian total error bar based criterion.}\label{nr2.1}
\end{figure}

We demonstrate the model selection criteria using the example of the advection-diffusion equation (\ref{ad-1}, \ref{ad-2}). 
We calculate the three criteria respectively for all candidate models via the regularization path. The values are illustrated in Figure \ref{nr2.1}. Based on our simulation of the advection-diffusion equation, when the threshold is between 0.02 and 0.05, the tBGLSS algorithm could obtain the correct model. 

\paragraph{Mean squared error} In Figure \ref{nr2.1} (left), we observe that the squared error of the estimated coefficient starts from the smallest error with a full model (which includes all possible terms).
The error then goes up as the algorithm starts to enforce sparsity and goes down as the algorithm reaches the correct model. 
Finally, the squared error stays high with an incorrect sparsity pattern (some correct terms are eliminated) after threshold=0.06. 
Even if this evaluation metric already utilizes practically impossible information (the true coefficient), this evaluation is certainly suboptimal. 
This is because when threshold=0, the least-square solution is the best unbiased linear estimator (BLUE), which leads to the best MSE. 
Therefore, this evaluation metric will always select the full model as the optimal choice, which certainly fails to provide the correct sparsity pattern for equation discovery. 

\paragraph{AIC-like loss} In Figure \ref{nr2.1} (middle), the AIC-like loss is large for the full model when threshold=0. Then, the error gradually decreases when approaching the correct model when the threshold is between 0.02 and 0.05. When the threshold is greater than 0.06, the AIC-like turns up slightly as some of the correct terms are eliminated by the algorithm. 
Compared to the MSE, the AIC-like loss is more acceptable as the correct model has the smallest loss. However, the candidate models collapse in the loss landscape. It is not clear enough to distinguish the correct model from the model with some minor terms missing (when threshold=0.06). From Occam's razor, one should always select the sparser model under similar performances. Therefore, the AIC-like loss could also be problematic. 

\paragraph{Bayesian total error bar} Here, the Bayesian total error bar provides much better results compared to the MSE and the AIC-like loss. 
In Figure \ref{nr2.1} (right), when the model approaches the correct sparsity pattern, the error drops sharply with the threshold between 0.02 and 0.05 and jumps back quickly as threshold=0.06. 
Unlike the MSE, the Bayesian total error bar clearly suggests the model with the correct sparsity pattern. 
Additionally, the error of the optimal model is significantly smaller than all other models, which magnifies the result from the AIC-loss. 
This is because when one (or more) correct term(s) is/are eliminated from the model, the level of unexplained variance will be large. 
Therefore, the posterior distribution will be less concentrated, which will result in large values for the total error bar. 
In summary, the Bayesian total error bar metric greatly outperforms other model selection criteria for sparse model discovery and PDE identification purposes.

\subsection{Dealing with Large Noise} \label{sec:largenoise}
It is hard to identify the correct partial differential equation when the noise is large, as we have shown in the former section. A subsampling method has been proposed in \cite{zhang2019robust}, but the time cost is huge with non-constant coefficients. Another way to overcome the problem is to reduce noise in the data preprocessing step. There are various ways to do this, and in our paper, we only consider three of them: moving average, Savitzky-Golay filter \cite{savitzky1964smoothing} and \textit{filtfilt} in \textit{SciPy}. We use the example of Burger's equation in (\ref{burgers-1}, \ref{burgers-2}), with 5\% noise added, to compare the three methods.

The mean squared error of the processed data to the original clean data can be used for the evaluation of the filter's ability:
\begin{equation}
    \text{MSE of $u'$}=\frac{\sum_{i=1}^n\sum_{j=1}^m[u'(x_i,t_j)-u(x_i,t_j)]^2}{mn}
\end{equation}
The data we used with 5\% white noise added have a mean squared error of $8.099\times 10^{-5}$.

\begin{figure}[H]\centering
  \includegraphics[width=7cm]{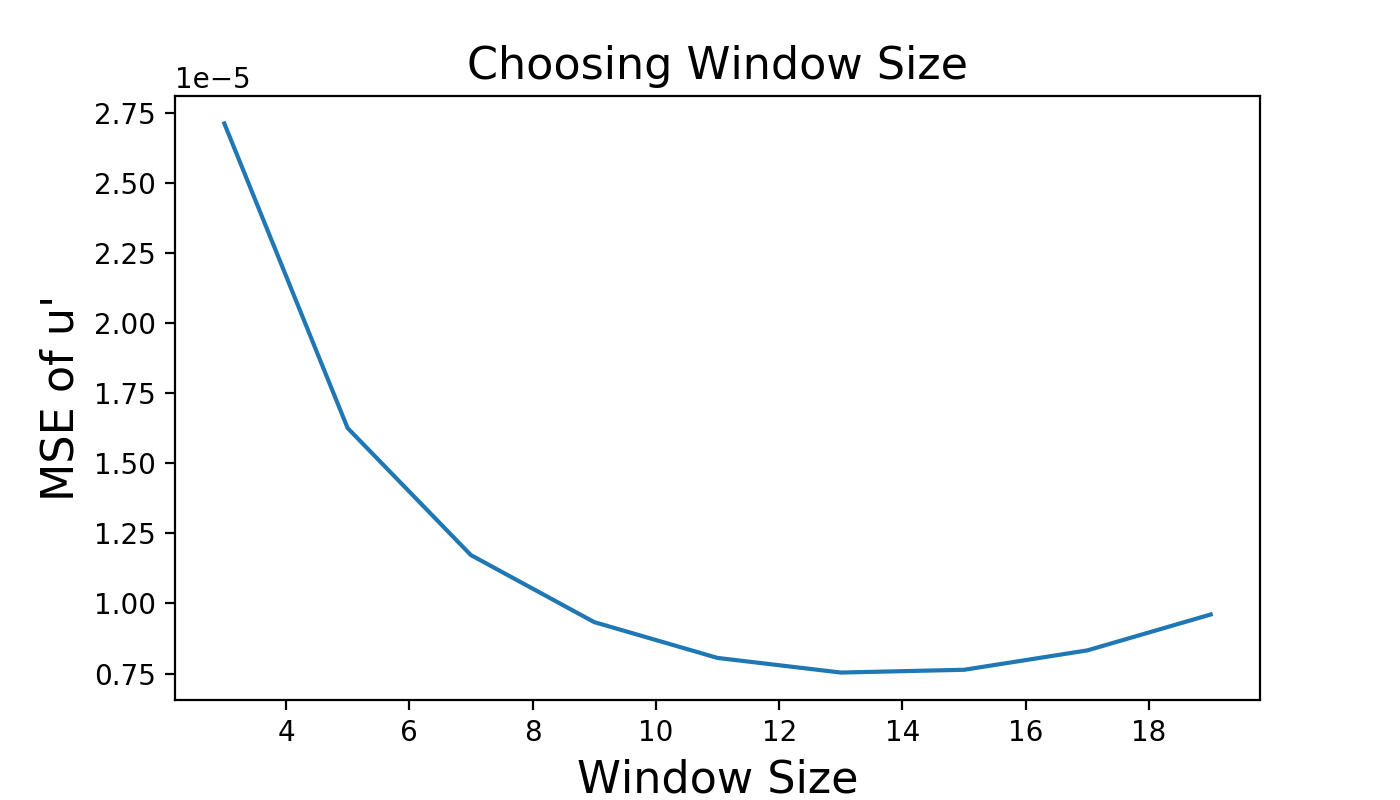}
  \caption{Nine choices of moving average windows and their mean squared errors of processed data. When window size is 13, MSE reaches minimum ($7.683\times10^{-6}$).}\label{nr3.1}
\end{figure}

\begin{figure}[H]
  \begin{minipage}{0.5\linewidth}\centering
    \includegraphics[width=7cm]{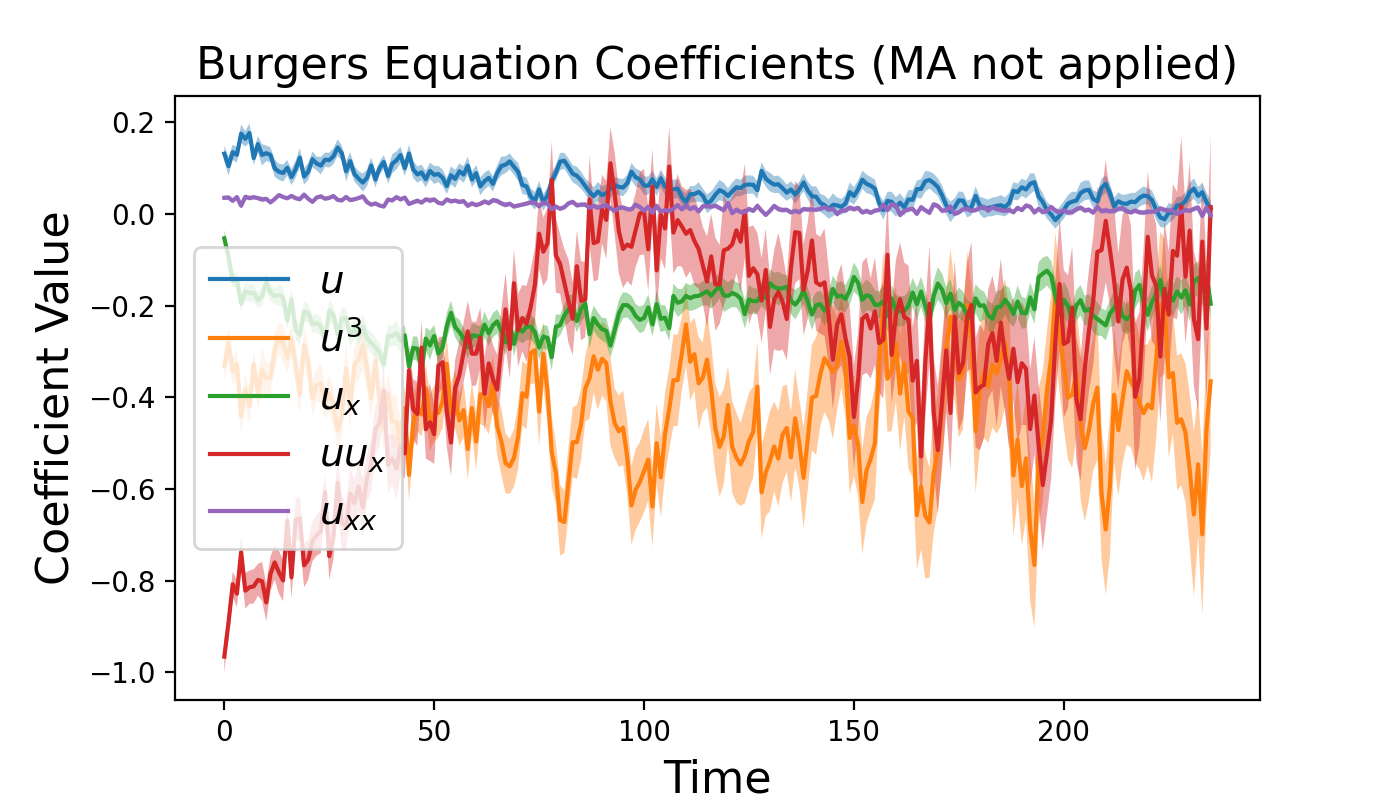}
  \end{minipage}
  \begin{minipage}{0.5\linewidth}\centering
    \includegraphics[width=7cm]{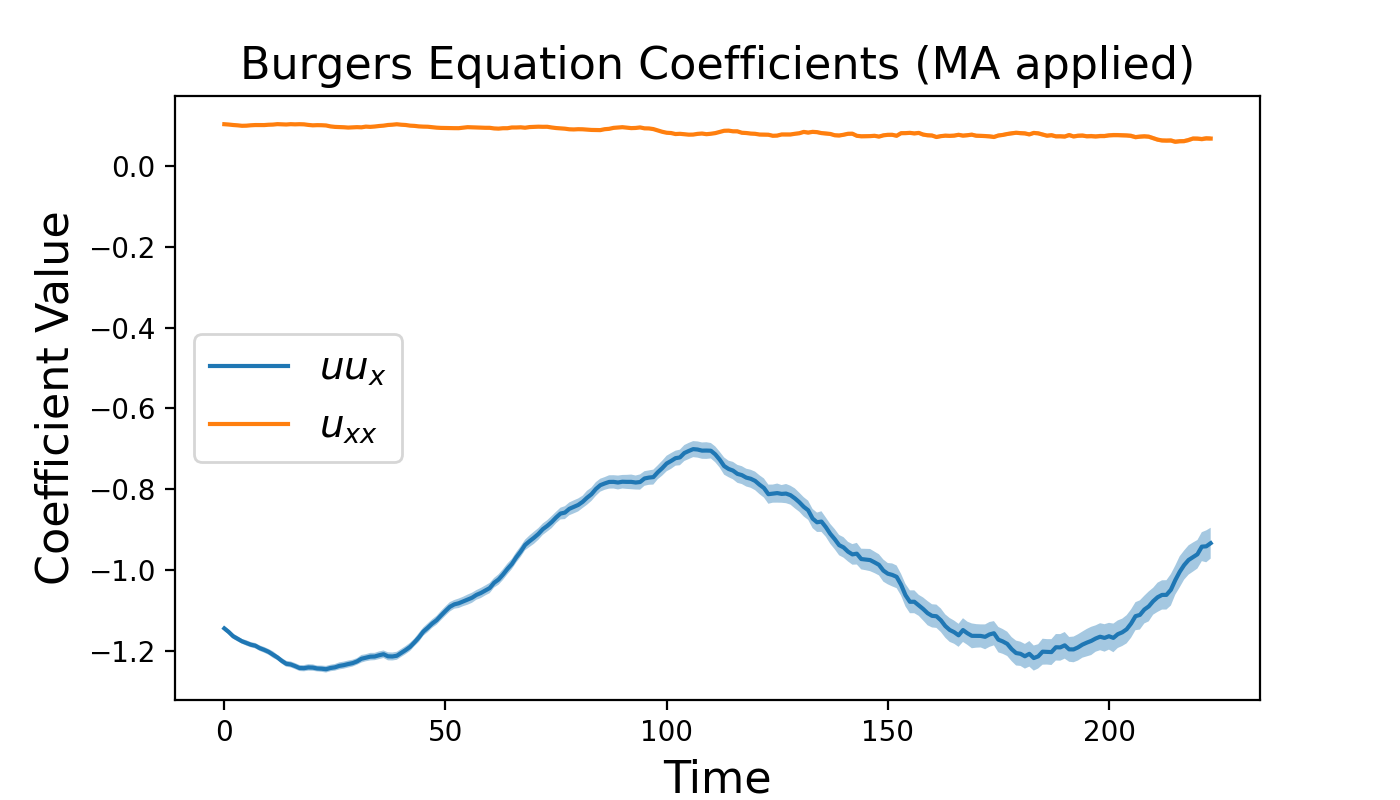}
  \end{minipage}
  \caption{Left: The coefficients identified in the Burger's equation example, with 5\% noise. Error bands for the coefficients are illustrated. Right: The coefficients identified from the same dataset, except for the noise preprocessed using moving average.  }\label{nr3.2}
\end{figure}

When using the moving average, the window size should be determined first. Draw the MSE of smoothed data against window size, we have the graph in Figure \ref{nr3.1}. The figure implies the best window width should be 13, where the MSE of processed data reaches its minimum ($7.683\times10^{-6}$). Feeding the smoothed data to our threshold BGL-SS algorithm, the coefficients can be correctly identified as shown in Figure \ref{nr3.2}, where the MSE of coefficients is $7.361\times 10^{-5}$. This is much better than that without the usage of moving average ($\text{MSE}=0.04244$). However, the moving average method has a few limitations. First, the edge of the dataset is cut. Second, it performs badly near the extremum as it tends to smooth them as well.

Savitzky-Golay filter can track the signal more closely than the moving average. Also, it will not change the shape of data. The filter also needs a window size to run. If the MSE of processed data is plotted against window size the same way, we can easily see that the MSE will reach the minimum ($6.504\times 10^{-6}$) when the window width is 37. Applying tBGL-SS on the smoothed data, the MSE of the coefficient is $7.363\times10^{-5}$, which is better than that of moving average. Notice that a large window size may distort the identification of coefficients.

The \textit{filtfilt} filter in \textit{SciPy} is a forward-backward filter. Instead of window size, the critical frequency in the lowpass Butterworth filter shall be set in advance. We discover that when the critical frequency is 0.0725, MSE reaches its minimum, $7.435\times 10^{-6}$. The MSE of coefficients given by tBGL-SS is $6.945\times10^{-5}$, which is at the same level as the Savitzky-Golay filter.

\section{Conclusions}
\label{conclusion}
In this paper, we propose a robust Bayesian sparse learning algorithm based on thresholding Bayesian group Lasso with the spike-and-slab prior (tBGL-SS) for PDE identification with spatial or temporal dependencies. 
Bayesian group Lasso with the spike-and-slab prior~\cite{xu2015bayesian} can ensure group sparsity of the posterior median estimators. We further speed up the computation from the idea of approximate MCMC by incorporating coefficient thresholds.
For uncertainty quantification, stepping forward from the extant approaches, tBGL-SS provides a distributional approximation of the PDE coefficients. It not only improves the performance of point estimation but also provides a theoretically grounded uncertainty estimation. Moreover, the quantified uncertainties could serve as a model selection criterion along the regularization path. 
Hence, we design and introduce the Bayesian total error bar criterion, which outperforms all classical sparse model selection metrics. 
Experimentally, we compare tBGL-SS to other baseline sparse grouped model selection methods~\citep{rudy2019data} using four examples to exemplify the power of tBGL-SS. 
The numerical results indicate that the tBGL-SS method is more robust than the baseline methods with large observation noises. 
It has a lower rate of false discovery compared to all other methods.

\enlargethispage{20pt}




\section*{Acknowledgement}

Guang Lin acknowledges the support of the National Science Foundation (DMS-2053746, DMS-2134209, ECCS-2328241, and OAC-2311848), and U.S. Department of Energy (DOE) Office of Science Advanced Scientific Computing Research program DE-SC0023161, and DOE–Fusion Energy Science, under grant number: DE-SC0024583. 
All data used in this manuscript are publicly available on \url{http://www.math.purdue.edu/~lin491/data/LETDC}. 



\vskip2pc

\section*{References}

\bibliographystyle{elsarticle-num}
\bibliography{main.bbl}

\begin{thebibliography}{10}
\expandafter\ifx\csname url\endcsname\relax
  \def\url#1{\texttt{#1}}\fi
\expandafter\ifx\csname urlprefix\endcsname\relax\def\urlprefix{URL }\fi
\expandafter\ifx\csname href\endcsname\relax
  \def\href#1#2{#2} \def\path#1{#1}\fi

\bibitem{koza1992genetic}
J.~R. Koza, J.~R. Koza, Genetic programming: on the programming of computers by
  means of natural selection, Vol.~1, MIT press, 1992.

\bibitem{bongard2007automated}
J.~Bongard, H.~Lipson, Automated reverse engineering of nonlinear dynamical
  systems, Proceedings of the National Academy of Sciences 104~(24) (2007)
  9943--9948.

\bibitem{schmidt2009distilling}
M.~Schmidt, H.~Lipson, Distilling free-form natural laws from experimental
  data, science 324~(5923) (2009) 81--85.

\bibitem{brunton2016discovering}
S.~L. Brunton, J.~L. Proctor, J.~N. Kutz, Discovering governing equations from
  data by sparse identification of nonlinear dynamical systems, Proceedings of
  the National Academy of Sciences 113~(15) (2016) 3932--3937.

\bibitem{mangan2016inferring}
N.~M. Mangan, S.~L. Brunton, J.~L. Proctor, J.~N. Kutz, Inferring biological
  networks by sparse identification of nonlinear dynamics, IEEE Transactions on
  Molecular, Biological and Multi-Scale Communications 2~(1) (2016) 52--63.

\bibitem{boninsegna2017sparse}
L.~Boninsegna, F.~N{\"u}ske, C.~Clementi, Sparse learning of stochastic dynamic
  equations, arXiv preprint arXiv:1712.02432.

\bibitem{tran2017exact}
G.~Tran, R.~Ward, Exact recovery of chaotic systems from highly corrupted data,
  Multiscale Modeling \& Simulation 15~(3) (2017) 1108--1129.

\bibitem{schaeffer2017bifurcation}
H.~Schaeffer, G.~Tran, R.~Ward, Learning dynamical systems and bifurcation via
  group sparsity, arXiv preprint arXiv:1709.01558.

\bibitem{schaeffer2017learning}
H.~Schaeffer, Learning partial differential equations via data discovery and
  sparse optimization, in: Proc. R. Soc. A, Vol. 473, The Royal Society, 2017,
  p. 20160446.

\bibitem{schaeffer2017sparse}
H.~Schaeffer, S.~G. McCalla, Sparse model selection via integral terms,
  Physical Review E 96~(2) (2017) 023302.

\bibitem{schaeffer2017extracting}
H.~Schaeffer, G.~Tran, R.~Ward, Extracting sparse high-dimensional dynamics
  from limited data, arXiv preprint arXiv:1707.08528.

\bibitem{mangan2017model}
N.~M. Mangan, J.~N. Kutz, S.~L. Brunton, J.~L. Proctor, Model selection for
  dynamical systems via sparse regression and information criteria, Proc. R.
  Soc. A 473~(2204) (2017) 20170009.

\bibitem{kaiser2017sparse}
E.~Kaiser, J.~N. Kutz, S.~L. Brunton, Sparse identification of nonlinear
  dynamics for model predictive control in the low-data limit, arXiv preprint
  arXiv:1711.05501.

\bibitem{loiseau2018constrained}
J.-C. Loiseau, S.~L. Brunton, Constrained sparse galerkin regression, Journal
  of Fluid Mechanics 838 (2018) 42--67.

\bibitem{rudy2017data}
S.~H. Rudy, S.~L. Brunton, J.~L. Proctor, J.~N. Kutz, Data-driven discovery of
  partial differential equations, Science Advances 3~(4) (2017) e1602614.

\bibitem{zhang2018robust}
S.~Zhang, G.~Lin, Robust data-driven discovery of governing physical laws with
  error bars, Proceedings of the Royal Society A: Mathematical, Physical and
  Engineering Sciences 474~(2217) (2018) 20180305.

\bibitem{zhang2019robust}
S.~Zhang, G.~Lin, Robust data-driven discovery of governing physical laws using
  a new subsampling-based sparse bayesian method to tackle four challenges
  (large noise, outliers, data integration, and extrapolation), arXiv preprint
  arXiv:1907.07788.

\bibitem{rudy2019data}
S.~Rudy, A.~Alla, S.~L. Brunton, J.~N. Kutz, Data-driven identification of
  parametric partial differential equations, SIAM Journal on Applied Dynamical
  Systems 18~(2) (2019) 643--660.

\bibitem{yuan2006model}
M.~Yuan, Y.~Lin, Model selection and estimation in regression with grouped
  variables, Journal of the Royal Statistical Society: Series B (Statistical
  Methodology) 68~(1) (2006) 49--67.

\bibitem{friedman2010note}
J.~Friedman, T.~Hastie, R.~Tibshirani, A note on the group lasso and a sparse
  group lasso, arXiv preprint arXiv:1001.0736.

\bibitem{kyung2010penalized}
M.~Kyung, J.~Gill, M.~Ghosh, G.~Casella, et~al., Penalized regression, standard
  errors, and bayesian lassos, Bayesian Analysis 5~(2) (2010) 369--411.

\bibitem{fung2023bayesian}
L.~Fung, U.~Fasel, M.~Juniper, Bayesian identification of nonlinear dynamics
  (bindy), Bulletin of the American Physical Society.

\bibitem{xu2015bayesian}
X.~Xu, M.~Ghosh, et~al., Bayesian variable selection and estimation for group
  lasso, Bayesian Analysis 10~(4) (2015) 909--936.

\bibitem{hobert1998geometric}
J.~P. Hobert, C.~J. Geyer, Geometric ergodicity of gibbs and block gibbs
  samplers for a hierarchical random effects model, Journal of Multivariate
  Analysis 67~(2) (1998) 414--430.

\bibitem{johndrow2020scalable}
J.~Johndrow, P.~Orenstein, A.~Bhattacharya, Scalable approximate mcmc
  algorithms for the horseshoe prior, Journal of Machine Learning Research
  21~(73).

\bibitem{newton1994approximate}
M.~A. Newton, A.~E. Raftery, Approximate bayesian inference with the weighted
  likelihood bootstrap, Journal of the Royal Statistical Society Series B:
  Statistical Methodology 56~(1) (1994) 3--26.

\bibitem{park2008bayesian}
T.~Park, G.~Casella, The bayesian lasso, Journal of the American Statistical
  Association 103~(482) (2008) 681--686.

\bibitem{savitzky1964smoothing}
A.~Savitzky, M.~J. Golay, Smoothing and differentiation of data by simplified
  least squares procedures., Analytical chemistry 36~(8) (1964) 1627--1639.

\end{thebibliography}

\end{document}